\documentclass[conference]{IEEEtran}

\usepackage{lipsum}     % package for generating dummy text
\usepackage{balance}    % balance the two columns of the paper in the last page
\usepackage{booktabs}   % professional tables. See http://cs.brown.edu/about/system/managed/latex/doc/booktabs.pdf
\usepackage{graphics}   % image files in figure (e.g., pdf, eps files)
\usepackage{url}        % formating a web url using
\usepackage{pifont}     % adding special characters using \ding{...}. See http://willbenton.com/wb-images/pifont.pdf
\usepackage[dvipsnames,table,xcdraw]{xcolor}
\usepackage[normalem]{ulem}
\usepackage{pgfgantt}
\usepackage{multicol}
\usepackage{changepage}
\usepackage{bbm}

\usepackage[utf8]{inputenc} % allow utf-8 input
\usepackage[T1]{fontenc}    % use 8-bit T1 fonts
\usepackage{hyperref}       % hyperlinks
\usepackage{url}            % simple URL typesetting
\usepackage{booktabs}       % professional-quality tables
\usepackage{amsfonts,amssymb}       % blackboard math symbols
\usepackage{nicefrac}       % compact symbols for 1/2, etc.
\usepackage{microtype}      % microtypography
\usepackage{lipsum}         % Lorem Ipsum fill text
\usepackage{multicol}       % Support for Multi columns for tables
\usepackage{multirow}       % Support for Multi rows fot tables
\usepackage{mathtools}      % Advanced mathtools
\usepackage{caption}        % Advanced caption configuration
\usepackage{amsmath}        % Math package for equations       
\usepackage{titlesec}       % Title section
\usepackage{graphicx}       % For adding labels to parts of the equation
\usepackage{stackrel}       % For adding labels to parts of the equation
\usepackage[ruled,vlined]{algorithm2e}  % For Algorithms
\usepackage{subcaption}
\usepackage{algpseudocode}
\usepackage{pgfgantt}
\usepackage{multicol}
\usepackage{changepage}
\usepackage{bm}
\usepackage{dblfloatfix}
\definecolor{tblue}{RGB}{93, 142, 150}

\newcommand{\cmnt}[1]{}
\definecolor{ForestGreen}{RGB}{34,139,34}

\DeclareMathOperator*{\argmax}{\arg\!\max}

\usepackage{stmaryrd}
\usepackage{trimclip}

\makeatletter
\DeclareRobustCommand{\shortto}{%
  \mathrel{\mathpalette\short@to\relax}%
}

\usepackage{bm}

\usepackage{amsmath,amssymb}

\author{\IEEEauthorblockN{Ramesh Doddaiah\textsuperscript{\textsection}}
\IEEEauthorblockA{
    \textit{Data Science}\\
    \textit{WPI}\\
    rdoddaiah@wpi.edu
    }
\and
\IEEEauthorblockN{Prathyush Parvatharaju\textsuperscript{\textsection}}
\IEEEauthorblockA{
    \textit{Data Science}\\
    \textit{WPI}\\
    psparvatharaju@wpi.edu
    }
\and
\IEEEauthorblockN{Elke Rundensteiner}
\IEEEauthorblockA{
    \textit{Data Science, Computer Science}\\
    \textit{WPI}\\
    rundenst@wpi.edu
    }
\and
\IEEEauthorblockN{Thomas Hartvigsen}
\IEEEauthorblockA{
    \textit{CSAIL} \\
    \textit{MIT}\\
    tomh@mit.edu
    }
}

\begin{document}

\title{Class-Specific Explainability for Deep Time Series Classifiers}
\maketitle
\begingroup\renewcommand\thefootnote{\textsection}
\footnotetext{Both authors contributed equally to this research.}

\thispagestyle{plain}
\pagestyle{plain}

%-----------------------------------------------
% Abstract
%-----------------------------------------------
\begin{abstract}
% For more detailed instructions, see here: http://dmkd.cs.wpi.edu/xkong/tutorials/blob/master/paper_writing/readme.md
% Post-hoc attribution methods highlights the time steps that are relevant for a specific time series classification by a neural network \cite{10.1145/3459637.3482446}. 
% Explainability methods help users trust machine learning models.
Explainability helps users trust
%high-quality 
deep learning solutions for time series classification.
% Recent works have  shown that explainability lowers barriers for users to adopt deep learning solutions for complex tasks, such as, time series classification.
However, existing explainability methods for multi-class time series classifiers focus on one class at a time, ignoring relationships between the classes.
% \ear{CHECK THe NEXT  REVISED SENTENCE - which has been revised -- DOES IT FIT YOUR STORY?}
 %\ear{
% THey therefore disregard critical semantics inherent in the multi-class scenario, where a classifier predicts the probabilities of multiple classes  by learning both  intra-class characteristics and inter-class differences.
%}
% relatively likelihood .
% We observe  that 
Instead, when a classifier is choosing between many classes, an effective explanation must show what sets the chosen class apart from the rest.
%In this work, 
We now formalize this notion, studying the open problem of class-specific explainability for deep time series classifiers, a challenging and impactful problem setting.
% In this context, class-specific explanations, essential in the multi-class scenario, have  been  overlooked.
% Meaning, the most-important time steps for a model's predicted class should be \textit{unique} to that class,  contrasting it  to  other classes.
% \ear{The method paragraph below seems to have
% no mention about if any of DEMUX is specific at all to time series -- it gives impression that it's all generic.}
% Don't split abstract into multiple paragraphs
% We are the first to address
% \textit{class-specific explainability} for deep time series classifiers, a challenging  real-world problem.
We design a novel explainability method, DEMUX, which learns saliency maps for explaining 
deep multi-class time series classifiers by adaptively ensuring that its explanation spotlights 
%only 
the regions in an input time series that a model uses specifically to its predicted class. %the distinguishing class characteristics relative to the explanations for the other classes.
DEMUX adopts a gradient-based approach composed of three interdependent modules that combine to generate consistent, class-specific saliency maps that remain faithful to the classifier's behavior yet are 
% \ear{This sentence 2nd part seems to be  a repeat instead of more deeply telling us the
% key strategies/ideas to be innovated by
% this paper -- what is time series specific;
% what specific things are class specific.
% we just talk about the goal and outcome, but no mention of how we succeed to do this -- the method. }
% that jointly 
% learn high-quality explanations that are class-specific
% , , consistent,
 easily understood by end users.
Our experimental study  demonstrates that DEMUX outperforms nine state-of-the-art alternatives on five popular datasets when explaining two types of deep time series classifiers.
Further, through a case study, we demonstrate that DEMUX's explanations indeed highlight what separates the predicted class from the others in the eyes of the classifier.

\end{abstract}

%-----------------------------------------------
% Introduction Section
%-------------------------------------------------------------------
% For more detailed instructions, see here: http://dmkd.cs.wpi.edu/xkong/tutorials/blob/master/paper_writing/readme.md

\section{Introduction}
\label{sec:intro}
\textbf{Background}.
Deep learning methods represent the state-of-the-art for tackling time series classification problems in important domains from finance \cite{Dixon2020FinancialFW} to healthcare \cite{Fulcher2017hctsaAC, Che2018RecurrentNN}.
In these applications, we typically aim to classify 
a time series to be  a member of one of many classes, 
referred to as  \textit{multi-class classification}
in contrast to binary classification \cite{Li2021ShapeNetAS}.
%
%For multi-class problems in particular, where 
% a time series to be  a member of one of many classes, deep learning methods are especially prevalent \cite{Li2021ShapeNetAS}.
%
% Time Series classification problem is one of the critical real-world problems in the time series data mining domain. 
% It has continuously received a significant amount of attention in recent years as they outperform most of the state-of-the-art traditional multi-class time series classification approaches. These deep learning methods are developed in a wide range of research domains and applications, such as sensors \cite{Mehdiyev2017TimeSC, Zhang2018HumanAR} and healthcare \cite{Fulcher2017hctsaAC, Che2018RecurrentNN}.
Despite the success of these deep learning methods,
%for classification, 
 domain experts may not  trust  predictions from deep models as they are ``opaque'' and thus hard to understand.
This lack of trust has the potential to hinder wide 
deployment of promising 
deep
models for real-world applications \cite{tonekaboni2020went}. 
% Making deep models explainable by finding the evidence the model used increases its usefulness \cite{bento2020timeshap,mujkanovic2020timexplain}. 
By helping users better understand and thus trust their models, \textit{explainability} has been
recognized as a critical 
% have become a popular and effective 
tool for successfully deploying deep learning models \cite{Schlegel2021TSMULELI,Parvatharaju2021LearningSM,Crabbe2021ExplainingTS,bento2020timeshap,Guidotti2020ExplainingAT,Guillem2019AgnosticLE,Ribeiro2016WhySI,NIPS2017_7062}. 
% \tnote{these are pretty old citations---grab some new stuff please. And add more!}
% To help users trust their models, \textit{explaining} a model's predictions is highly beneficial \cite{Ribeiro2016WhySI}. %, recent works have begun making them explainable by highlighting which input time steps the model found most useful \cite{bento2020timeshap,mujkanovic2020timexplain}.
% \ear{DO YOU MEAN MULTI-LABEL OR
% MULTI-CLASS HERE?}
% \ear{AND MULTI-CLASS VS WHAT? WHAT DID OTHERS DO?}

To  explain a time series classification,
we  highlight the time steps that the model associates with the class it  predicts via a   saliency map 
\cite{ismail2020benchmarking,bento2020timeshap,Parvatharaju2021LearningSM}.
However, 
in the multi-class 
problem setting where a classifier is choosing between many classes, an effective explanation must show 
\textit{what sets the chosen class apart from the rest of the  classes}. Thus the explanation should only 
highlight  that particular subset of time steps that explains
why that class was predicted compared to other classes.
% \textit{given multiple classes to choose from}.
%  to effectively explain a  classification in the multi-class  problem setting, 
%  we should be  highlighting \textit{only} the
% subset of time steps that the model associates with its predicted % class.
%
% Otherwise, the  explanation will not describe why that class was predicted \textit{given multiple classes to choose from}.
% Intuitively, to explain a multi-class prediction, an explainability method must determine \textit{what set this class apart from the rest?}
To further illustrate this  with an example from computer vision, consider classifying images as either cats, dogs, foxes, or wolves. Fur is certainly evidence that an \textit{animal} is in the image, but given choices between only \textit{these} animals, fur is \textit{specific} to no class.
Therefore, an explanation should avoid highlighting fur \cite{Shimoda2016DistinctCS}. In computer vision, it has been recognized
as critical to leverage 
relationships  between classes  \cite{Shimoda2016DistinctCS}.
We conjecture that the same is equally true for time series, yet existing methods for time series ignore all relationships between classes \cite{Schlegel2021TSMULELI,bento2020timeshap,Crabbe2021ExplainingTS,Guidotti2020ExplainingAT,Ribeiro2016WhySI}.
% unlike in computer vision \cite{Shimoda2016DistinctCS}.
% \ear{could we cite some papers here, without offending anyone?
% lime ! other such popular methods !}
% \ear{ABOVE, THERE IS SOME AMBIGUITY HERE - HAS MULTI-CLASS BEEN
% SOLVED IN GENERAL INCLUDING THE ABOVE OBSERVATION ABOUT CATS
% AND DOGS? AND ONLY THE FACT THAT THIS IS "ALSO" TRUE WHEN
% THE DATA TYPE IF TIME SERIES IS WHAT IS NEW HERE? OR WHAT
% IS OUR CLAIM !!?  WHAT IS STATE-OF-ART HERE, RE WHAT CLAIM WE STATE HERE.}
% \EAR{ALSO, I'M SLIGHTLY CONCERNED THAT THE EXAMPLE WE PICK TO EXPLAIN
% OUR ABOVE POINT/S GIVES THE FEELING OF A BINARY EXAMPLE OF CAT VS DOG. CAN WE MAKE THIS GO AWAY BY STARTING WITH SAYING SOMETHING ABOUT CLASSIFICATION AMONG MANY ANIMALS...?}
Further, by ignoring this class-specificity issue, users may be prone to rationalize a model's prediction in the face of erroneous explanations, having been instilled with a false sense of confidence \cite{Kaur2020InterpretingIU}.

\begin{figure}[t]
\centering
  \includegraphics[scale=0.53]{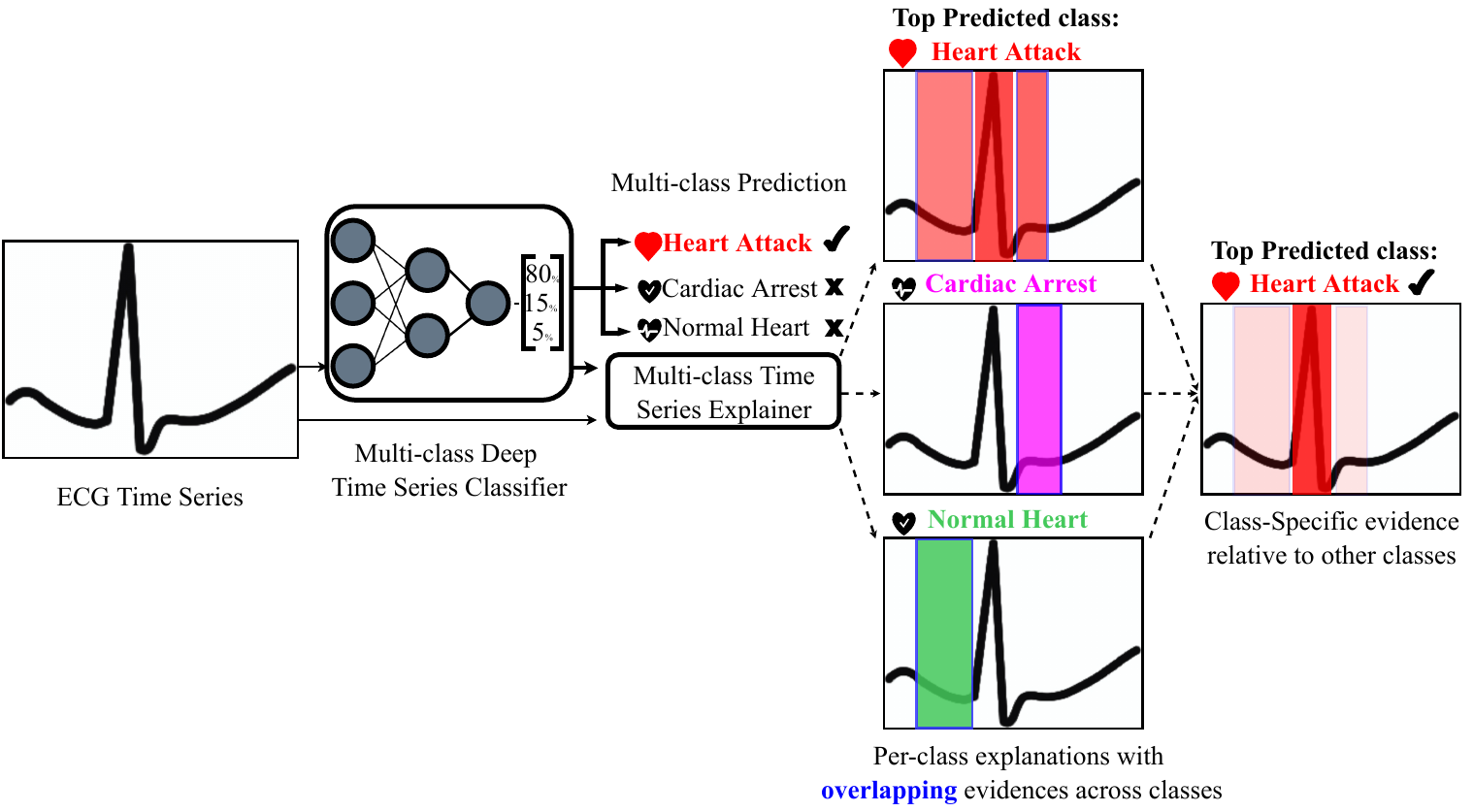}
  \caption{
  Effective explanations of multi-class time series classifiers should show what regions of a time series set that class apart from the rest.
  Here, we show a deep classifier that predicts \textit{Heart Attack} given an ECG.
  Explanations for individual classes (middle column) may have overlapping regions. However, as shown on  right, the middle of the signal is what sets \textit{Heart Attack} apart according to the classifier.
%   This shows class-specific important regions (the rightmost image) contributing to classification results. 
%   The classifier shows low confidence due to overlapping evidences as shown in the blue rectangles across per-class explanations.  
%   %   \snote{This says that preserving overlapping region is importance as it is the evidence for confidence score. Probably defeats our purpose of removing the overlapping regions? The last sentence of Motivating example section says the opposite} 
%   The model's interpretability can be improved with class-specific explanations as shown in the rightmost image after learning to reduce overlapping regions from the top predicted class.  
  %   \snote{Cardiac Prediction -> Cardiac Arrest, Normal Prediction -> Normal}
  %  \ear{It is not clear why this type of explanation  you propose is
%  useful!?  And, where in the visual we can see the class specific explanation? point us, top right, middle? Also,  you show us 2 saliency maps for the heart attack case --
%  some more text here would be helpful. or you can use the white space inside the figure for some annotations.}
  } 
  \label{fig:MotivatingExample}
%   Example of distinct evidence for the top predicted class highlighting class-specific salient regions contributing to classification results.
% \vspace{-8mm} \tnote{don't put this vspace modifier in}
\end{figure}

% \ear{in fig 1,  it talks about  60 percent confident, but in
% the example desc. below on notion of confidence is mentioned?
% is that mismatch on purpose?}

%fixed the font size in fig 1
% \ear{CHECK IF THE 
% WORDS in fig 1 are
% big enough to comfortably
% read without enlarging.
% it looks rather small font now.
% Did you shrink the figure - not a good idea ...}

\textbf{Motivating Example.}
Consider a deep multi-class time series classifier for ECG data \cite{Ribeiro_2020} that classifies heartbeats into three classes: \textit{Heart Attack}, \textit{Cardiac Arrest}, and \textit{Normal Heart}.
Without knowing whether the classifier learned to recognize relevant critical regions of the time series for each class, a doctor may not trust its prediction. As shown in Figure \ref{fig:MotivatingExample} (middle), state-of-the-art methods would explain 
each class  independently \cite{Crabbe2021ExplainingTS,Parvatharaju2021LearningSM,NIPS2017_7062,Ribeiro2016WhySI}.
% \ear{CITE SOMETHING HERE \cite{XXX}}
%
%each class could be explained independently. 
%
However, by ignoring the relationships between classes, 
it appears  as if the model uses 
% \ear{STILL careful -  because the red saliency map
% does not cover the whole signal/time series, only a subset of it}
most of the signal when predicting its top class, \textit{Heart Attack}, even when it shares the same critical regions with Cardiac Arrest and Normal Heart classes as shown in blue. A class-specific explanation, on the other hand, 
should correct for the model using the same regions for multiple classes. 
% \ear{(e.g., Heart Attack and Cardiac Arrest share
% significant common evidence, and so do Heart Attack and Normal Heart).}.
In our example, it (rightmost column) explains that the burst (non-overlapping critical region) of the signal is what the model uses to predict  \textit{Heart Attack} instead of  other classes.
% \ear{Above is a  vague, as to if 
% we care about if the relevant signal is
% in a certain position, middle, right, left, or if it
% looks like a certain shape?
% maybe that is a side-point here - still thinking}

% shown on the right, highlights only the region of the input that uniquely contributes to the model's 80\% confidence of \textit{Heart Attack} class (in contrast to any other classes). % brings more clarity to the end user and builds trust in the deep learning models.

% these per class saliency maps by existing methods show overlapping salient regions as shown in the blue color. Figure \ref{fig:MotivatingExample} bottom row shows unique evidence without any overlapping salient regions bringing much needed clarity for the end users. 
% that is relative to other classes. 

% \tnote{we want to make sure we're not discussing a solution here: The goal of this figure should be to illustrate what it looks like to do a good job on this problem: Outputs should be more clearly "distinct". In this version it is hard to see that middle explanations are not distinct but the one on the right is distinct. Can we point to the overlap or lack of overlaps? Another version of this figure could be something like Figure 1 in my KDD'20 paper: Show examples of good and bad solutions.}. 

\textbf{State-of-the-Art.}
% Recently, a significant amount of work is going on to elucidate the key patterns 
% \snote{key features? use of patterns is much more general} learnt by a deep learning opaque-box model \snote{We can jump ahead and focus on time series nature instead of describing the general pattern}. 
%
Explainablility for time series models 
has recently emerged as a promising direction to help users trust deep time series classifier models
\cite{Crabbe2021ExplainingTS,bento2020timeshap,Parvatharaju2021LearningSM,Schlegel2021TSMULELI,Guidotti2020ExplainingAT,mujkanovic2020timexplain, Ribeiro2016WhySI}.
%
%
% Interest in explainablility for time series models is on the rise \cite{Crabbe2021ExplainingTS,bento2020timeshap,Parvatharaju2021LearningSM,Schlegel2021TSMULELI,Guidotti2020ExplainingAT,mujkanovic2020timexplain, Ribeiro2016WhySI}, and has emerged as a promising direction to help users trust deep time series classifier models. 
%
The most successful methods \textit{learn to perturb} input time series to explain an opaque model's behavior 
% \ear{CAN YOU EXPLAIN MORE WHY INTUITIVELY
% THAT EXPLAINS ANYTHING TO LOOK AT VICINITY}
in the vicinity of one instance \cite{Parvatharaju2021LearningSM, Crabbe2021ExplainingTS}. Intuitively, time steps that have a higher impact on model accuracy will be ranked higher.
% \ear{what does "near" mean here?} near to the input time series.
% \ear{you need to add a CITATION here, as you are referring
% to successful models, yet you list none.}
%
%
Most existing methods \cite{Crabbe2021ExplainingTS,bento2020timeshap, mujkanovic2020timexplain, Ribeiro2016WhySI,NIPS2017_7062,tonekaboni2020went} explain model behavior by perturbing each time step using either static, predefined values like zero or
other time series instances from a ``background'' dataset.
For example, PERT \cite{Parvatharaju2021LearningSM}, which explains only binary deep time series classifiers,  perturbs each time step by replacing it from a replacement time series 
sampled 
from the background dataset. 
% \ear{Next sentence is not sufficient --
% so it says that this method handles
% multi-class problem -- you only say that it uses
% perturbation approach - but you forgot to mention
% if it directly takes on the class-specificity
% in the context of the multi-class issue or if it ignores
% this and treats all classes as if they were indept from other classes.}
DYNAMASK \cite{Crabbe2021ExplainingTS,Fong2017InterpretableEO}, which also treats each class independently, uses static replacement strategies for each time step for deriving explanations for a multivariate classifier. It makes a binary decision if a feature is important or not. To-date, class-specific explanations, despite their
recognized need in  fields like computer vision \cite{Shimoda2016DistinctCS}, remain an open problem in time series.
Typically, a successful multi-class classifier assigns high probability to one of the classes and lower probabilities to the rest. Evidence derived to explain the predicted class should be unique to that class, relative to other classes.
But existing time series explainable methods
% \ear{?? all existing methods? or those
% focused on time series?? but do the
% ones focused on images  do this?}
fail to  incorporate the knowledge about
%acknowledge 
relationships between classes.

Beyond lacking class-specificity, another well-known disadvantage of perturbation-learning methods is the high variance between explanations  derived over multiple runs for the same time series instance as input
\cite{Kumar2020ProblemsWS,Fong2017InterpretableEO,fong2019understanding}.
High variability among explanations decreases a user's trust in an explainability method and must therefore be reduced.

\textbf{Problem Definition.} 
We study the open problem of \textit{Class}\textit{-Specific} \textit{Explainability for Multi-Class Time Series Classifiers}: given a time series  and a pre-trained multi-class classifier, we aim to generate a 
\textit{class-specific saliency map} for the classifier's 
predicted class.
% highest-probability class.
A saliency map is a vector with one element per time step in the time series instance, where higher values of an element  indicates a higher importance
of this time step according to the classifier.
% For each element, higher values indicate higher importance for the corresponding time step, while lower values indicate lower importance.
To be \textit{class-specific}, the saliency map should assign high importance only to time steps  uniquely important to the predicted class (in contrast to also being important to other classes).
% Our goal is to generate a saliency map for the top predicted class.
% A vector where higher values indicate stronger dependence of the model on the unique time step for the class. A unique time step explains only the top predicted class and is not shared with other classes.
% and remove overlapping salient regions across the classes from the top predicted class to produce a distinct \tnote{reader has never heard of a "distinct" saliency map} saliency map \tnote{this just sounds like your method... not the problem! The problem is nothing to do with HOW you do it.. just what you want to end up with: A saliency map for the top class for which the saliency values are unique. Then define unique}.
This problem has multiple possibly conflicting objectives: a good saliency map should be class-specific, highlight only the most-relevant time steps, and still remain faithful to the model's behavior.

\textbf{Challenges.}
Our problem is challenging for several  reasons:
\begin{itemize}
% Generating explanations with shared salient regions, is like explaining intermediate classes. 
% Explaining shared salient regions
%
% To avoid generating explanations explaining these overlapping regions 
% To discriminate the salient regions of a target predicted class from the the regions of the other classes
% We don't want to generate explanation Considering the explanations of other classes and their impact on the model's predictions.
%
%Maybe we don't need to prefix with "Learning" word
\item \textit{Class-Specificity}: Generating class-specific saliency maps requires knowledge of explanations across all classes. 
However, learning concurrently multiple explanations 
is hard, in particular for low-probability classes,
with a model's predictions  often highly variable in regions of low probability.
\item \textit{Local Fidelity}: We consider multi-class classifiers that predict probability distributions. Learning perturbations to explain these models must incorporate all class probabilities  to remain faithful to the classifier's behavior. However, minor changes to the input can have a large effect on the predicted class distribution. % Generated perturbation-based class-specific explanation needs to preserve this probability distribution to obtain an explanation that is locally faithful to the classifier.
% This requires the need to learn class-specific data replacement and use in-distribution perturbations.  
% \item \textit{Local Fidelity}: multi-class classifiers predict probability distribution across multiple classes. Generated class-specific explanation needs to maintain this prediction probability distribution to obtain an explanation that is locally faithful to the classifier. We need to learn per-class saliency maps jointly to preserve the predicted probability distribution and learn to perturb each class with in-distribution and class-specific data replacement to produce locally faithful explanations. \tnote{last sentence is too long and drawn out. should be more concise!}
% Learning to remove shared saliency across the classes necessitates the need to learn per-class saliency maps jointly. 
% \tnote{I sort of see what you're saying, but somehow it is very unclear. Maybe too much jargon? Makes it too hard to follow}
% \snote{Not motivating enough to learn saliency maps jointly}

\item \textit{Temporal Coherence}: Time steps often depend on their neighbors' values.
This implies that similarly
 for saliency maps  neighboring time steps should have similar importance.
 %
%The same should be true for saliency maps in that neighboring time steps should have similar importance.
While this encourages discovering important subsequences, thereby improving explainability, it conflicts with local fidelity and class-specificity. Hence, a trade-off  must  inherently be considered
in any effective solution.% improves explainability by encourage a model to find important subsequences, Neighboring time steps should have similar impacts on the predictions of the model because they are often strongly correlated. An understandable saliency map should have contiguous important regions without sacrificing class-specificity. These two goals naturally conflict, requiring a trade-off.

% I need a better word here, instead of Consistency
\item \textit{Consistent Saliency:} Perturbation-based explainability methods can create saliency maps that vary dramatically for the same instances when 
%they are
re-initialized. 
% create high variability between explanations for the same time series instance across multiple runs.
Yet to be useful in real-world applications,
%perturbation-based methods 
we should instead consistently generate similar explanations for the same time series

%PP and RD decided to keep this Consistent Saliency paragraph.
% \tnote{does our method really do this? And do our experiments prove it? If not, remove this. Looks like the AUCs down below have very low standard deviation, so this claim doesn't seem to be backed up?}.

% At the same time, we do not need overlapping saliency across the classes for these same time steps \tnote{i'm not sure what this means}. 

% \item \textit{multi-class Replacement Time Series}:
% \item \textit{Class-Specific Perturbations}: Post-hoc explanations are generated by perturbing each time steps with in-distribution data

% \item \textit{Learning Cluster Specific Replacement Time Series} \tnote{first mention of clustering? Remember these are challenges of the PROBLEM, not our solution. The reader doesn't care if we chose to solve it in a way that was hard. That makes a worse story.}: Time series within the same class and across the classes can be highly variable even within the dataset. Perturbation of time series needs to be done using in-distribution data. So, we need to find a suitable cluster \tnote{no---that's just how we solve it} of sample replacement time series per-class from the background dataset and learn a perturbation strategy that adapts to these replacement time series. \tnote{too close to our method}

%RD: TODO: We need to choose 2 or 3 challenges here
\end{itemize}

% \tnote{in the state-of-the-art, you say that there's high variance between explanations, then never mention it again!!! I thought that would be a challenge and part of the solution. If it's not, then remove that paragraph above}
% \tnote{we can do 3 sharper challenges. Can we recycle any good ones from last time? First challenge should be for Class-Specificity (it is already). Second challenge should be XAI for time series specifically is hard. Third challenge should be the temporal coherence one, that is good.}
%Needs more work here

%Temporal perturbation by prioritized replacement for multi-class
%Class-Specific Distinct Temporal Explainer

\textbf{Proposed Solution.} 
To derive class-specific explanations, 
% attribution-based 
% \tnote{this is the first time you say "attribution-based" in the intro. Don't introduce new language late like this, keep leaning on the same language we use above!} 
we propose \textit{\textbf{D}istinct} \textit{T\textbf{E}mporal} \textit{\textbf{MU}lticlass} \textit{E\textbf{X}plainer (DEMUX)}, a novel model-agnostic, perturbation-based  explainability method for multi-class time series models.
% \tnote{I'm not a fan of LMUX, it is too clunky. MUX is ok but nothing to do with time series. Could add "temporal" somewhere? or perturbation? we need to brainstorm for this a bit}.
% DEMUX is a gradient-based learning algorithm 
% % \tnote{does the algorithm resemble a "search algorithm"? That's kind of loaded language}
% to discover the impact of perturbing each time step with varying degrees across the classes on the opaque box model's predicted probabilistic distribution. It produces a learned class-specific saliency maps 
% \tnote{break into two focused sentences}. 
%
% below when "then" -- do we need to know about this
% seq of steps here.
%
%DEMUX jointly learns saliency maps for all classes, then learns to remove shared salient regions to generate a class-specific explanation for the model's predicted class.
%
% To produce a class-specific saliency map for the classifier's top predicted class, DEMUX estimates maps for each class along the way.
DEMUX jointly learns saliency maps,
% for all classes
%% EAR: i removed -- "for all classes" - are the (n-1) other maps not class-specific?
%\ear{IS IT FOR ALL, or for the predicted class -- or do you want to leave it vague here and remote the word "all" or remove the words "for all classes"? THIS IS CONFUSING. SO IS THIS DONE FOR ALL  CLASSES OR ONLY FOR THE TOP CLASS THIS SPECIAL TREATMENT OF DIFFERENTIAL EXPLANATION?} \tnote{I think this is confusing because the order of the story is rather odd.. instead of linearly building towards a complete system, we're jumping in and out of different components at different times}
% all classes,
with a focus on  removing shared salient regions to generate a class-specific explanation 
for the model's top predicted class.

DEMUX is gradient-based approach that monitors changes in the classifier's predictions while perturbing values at each time step. %discovering the impact of perturbing each time step on the classifier's predictions. 
% \ear{HERE IS SEEMS TO TALK ABOUT ONLY PRODUCING
% ONE MAP, and not one for each class????
% inconsistent with the problem statement!}
It produces a saliency map for the classifier's top predicted class that preserves the classifier's prediction probability distribution across classes. To generate good perturbations, DEMUX learns to sample a replacement time-series per class from a background dataset 
%
% containing many time series 
%
using a clustering-based replacement selector. DEMUX avoids out-of-distribution replacement values by ensuring perturbations are like other time series the model has seen before for each class and for each time step, leading to more stable saliency maps. 

% LMUX jointly learns a specific function \tnote{vague language, should try to make it sound more lively} to identify and remove shared saliency between the target class and other classes \tnote{this is an important sentence that is hidden. It should be the very first thing: Jointly learns saliency maps for all classes and then integrates them to remove shared regions. That still sounds a little dull to me :)}. 

% Using a \textit{kNN cluster-based prioritized replacement selector} we learn to sample a replacement time-series from an optimal \tnote{we can't say it's actually optimal} cluster of time series instances that is built by available background dataset. This way it adapts to the given dataset, and to each time step of the instance-of-interest. \tnote{not super clear how everything relates to the challenges or what exactly happens to make sure that the outputs achieve our primary goal of "class-specific"}
% % \snote{Jointly learn with what?}

% doubt if we can use this new word https://antoinesavine.com/2020/05/04/differential-machine-learning/

% \ear{I AM CONCERNED THAT WE DO NOT CIRCLE
% BACK TO THE 4 TECHNICAL CHALLENGES EARLIER,
% AND MAKE ANY STATEMENTA ABOUT IF WE ADDRESS
% ANY PARTICULAR OR EVEN ALL 4 OF THE
% STATED CHALLENGES. WE SHOULD SAY SO
% EXPLICITLY IN OUR BRIEF APPROACH
% PARAGRAPH LISTING THE CHALLENGE OVERCOME BY ITS NAME HERE....}

\textbf{Contributions.} Our main contributions are as follows:
\begin{itemize}
\item We identify and characterize the problem of class-specific saliency maps for deep multi-class time series classifiers.
\item We introduce the first effective solution, DEMUX, which extends beyond recent work with three innovations: learning to remove shared saliency across classes (Class-Specificity), generating class-specific perturbations that are locally faithful (Local Fidelity and Temporal Coherence), and ensuring stability of saliency maps (Consistent Saliency) for given time series instances.
% \ear{in abstract you said there were 3 components.
% maybe one component is boring and thus not listed here?
% or better, we can just use another term here --  i'll plug in 2 innovations!?}
% First, DEMUX learns to remove any shared saliency from the predicted class to highlight class-specific salient regions that contribute to the prediction. Second, DEMUX learns class-specific data replacements to generate locally faithful explanations. Third, DEMUX learns to generate consistent saliency maps with low variance between explanations for the same input time series instances.

% A common problem with perturbation based methods [ 13 ],
% [11], [ 16] is high variance between explanations, even given
% the same inputs multiple times.

% Lastly, DEMUX consistently generates similar explanations with less variation for the same time series instance across multiple runs. \tnote{we do not need to say how it works here.. that's in the last paragraph. I thought I removed this?}
\item Using five real datasets, we conclusively demonstrate that DEMUX outperforms nine state-of-the-art alternatives, successfully generating class-specific explanations
%%% added below - does it at least hint at model agnostic claim? maybe too subtle?
for
multiple types of deep time-series
%opaque 
classifiers.
% \ear{We used "big words" earlier like DEMUX
% being model-agnostic - and we never circled back to
% confirm/that that it actually is or that
% we show that it is}
\end{itemize}

%-----------------------------------------------
% Related Work Section
\section{Related Works}
Deep learning models have achieved  remarkable results in domains from time series forecasting to classification tasks. But these models are opaque in nature. Given that a substantial amount of data collected in high-stake domains like healthcare,
\cite{Fulcher2017hctsaAC, Che2018RecurrentNN} finance \cite{Kelany2020DeepLM} and autonomous vehicles \cite{Mehdiyev2017TimeSC, Zhang2018HumanAR}  are in the form of  time series, the need to build explainable AI (XAI) for the time series domain is rapidly increasing. 
\textit{Saliency maps} \cite{Fong2017InterpretableEO,fong2019understanding} are  among the many promising approaches to increasing transparency of a deep learning model. They 
correspond to 
%%
%%generate 
%% 
importance scores that highlight the  series regions contributing to the classifier's predictions. Below we  characterize XAI methods according to the strategy they utilize to  generate importance scores representing model explanations.

% for a model's prediction. 

\textbf{Perturbation methods}. These methods \cite{Crabbe2021ExplainingTS, Parvatharaju2021LearningSM, Fong2017InterpretableEO} slightly change the input time series and compare the output to the baseline to create an importance ranking.
 The resulting saliency  masks indicate importance of the  time steps, explaining prediction for the given test instance. In time series, each time step can have a range of valid or in-distribution values. Assuming static baselines replacement values or random time series from background datasets are good enough for each class-specific replacement tends to lead to sub-optimal importance scores for saliency maps.

\textbf{Surrogate models}. These methods \cite{Schlegel2021TSMULELI,Ribeiro2016WhySI,schlegel2019rigorous} produce linear surrogate models that are trained to emulate the behavior of an opaque model locally for a set of perturbed versions of a given instance. If the linear model is a good enough local approximation, then its coefficients are treated as an explanation. Intuitively, if there is not a clear linear relationship between the set of perturbed instances and the opaque model's predictions, which is often the case in complex time series deep learning models, a linear model's coefficients fall short of generating a good explanation for a non-linear model’s predictions.

\textbf{Other methods}. There are some popular methods \cite{bento2020timeshap,NIPS2017_7062,mujkanovic2020timexplain,Guillem2019AgnosticLE} that do not fall in one of the above categories. SHAP \cite{NIPS2017_7062} does random permutations of the input features and average the marginal contribution of features to build importance Shapley scores. TIMESHAP is built upon KernelSHAP and extends it to the time series domain.

% and its variants \cite{bento2020timeshap,mujkanovic2020timexplain,Guillem2019AgnosticLE} probes the model in the vicinity of the given test instance.
% By computing Shapley values for the  model's predictions, they compute a linear \ear{HOW DO YOU KNOW IT MUST BE LINEAR? WHAT DOES "PROBE" DO TO ASSURE THAT?  ALSO,
% WHY NOT STATE THIS EARLIER RE LINIARITY?
% if linear important as it indicates simplicity?
% }
% surrogate \ear{above paragraph was on surrogates? the
% way you wrote this paragraph there is not even any hit
% why you gave it the subject headine game theory, and
% with aspects of SHAP relate to or leverage game theory...}
% for the model's predictions. Shapley values provide the importance \ear{Is it plural? } score. These 
% \ear{What is different here between permutation and perturbation? stick with same terms, or elaborate how connects or even better how  intuitively how differs}
% permutation-based methods are
% \ear{why? put the point of "slow" last. slow compared to what?} slow 
% \ear{you did not mean in general! this seems unrelated
% to permutation, butit may relate to the papers you happen to cite above!?}
% and use random replacement strategies.

% \ear{We need a new bold title below; as 
% it's unclear if below refers to game theory specifically,
% or if you are stating below in general for all alternatives
% in the literature?}
% \ear{Gap in Methods.}
Unfortunately, none of the above category of methods address class-specific explanations challenges i.e., there has been little work developing explainability methods specifically for \textit{multi-class} setting in time series, despite the recognition of the class-specific explainability importance 
% \ear{Do you mean success 
% of multi-class models for classification, right,
% not for explanations?}
in computer vision domain \cite{Shimoda2016DistinctCS,Fong2017InterpretableEO}. Without leveraging knowledge from other classes during multi-class explanation learning, explanations from existing methods are less effective \cite{Shimoda2016DistinctCS}. To the best of our knowledge, our work is the first to study learnable class-specific perturbations to explain opaque multi-class time series classifiers. The methods mentioned above do not focus on generating \textit{class-specific} perturbations and consistent explanations. Nor do they preserve the opaque model's prediction probability distribution across the classes while generating explanations. 
%RD:Maybe we can remove last 2 statements in the above paragraph?

% \ear{BELOW, THIS SEEMS TO BE MORE A GENERAL PROPERTY DISCUSSION
% OF YOUR GOAL STATEMENT< MOTIVATING WHY YOU WANT TO DO THIS.
% BUT IT DOES NOT SEEM REQUIRED TO SAY SO in this RELATED WORK SECTION?}
% Generating a class-specific explanation 
% \ear{HM - so is that the aim  instead of for all classes???}
% for the top predicted class provides a higher level of abstraction for the end users. It thus avoids unnecessary confusion for visualizing the per-class saliency maps. Preserving a prediction probability distribution across the classes along with in-distribution, class-specific perturbation generates an explanation that is locally faithful to the classifier. 

% Users trust the explainability methods if it generates consistent explanations over multiple runs for the same input time series instance, but existing methods show high variability. In this work, we study this problem formally to generate consistent explanations.

%-----------------------------------------------
% % Problem Definition Section
% \input{sec_2_problem_def}

%-----------------------------------------------
% Method Section

%---------------------------------------------------------
\section{Methodology}
\label{sec:method}
\subsection{Problem Definition}
Assume we are given a set of $N$ time series $\mathcal{D} = \{\mathbf{x}_1, \dots,\mathbf{x}_N\}$ and a deep multi-class time series classifier $f:\mathbb{X} \to \mathcal{Y}$, where $\mathbb{X} \in \mathbb{R}^T$ is the $T$-dimensional feature space with and $\mathcal{Y} = \{1,2,\dots,C\}$  is the label space with $C$ classes. Let us consider an instance-of-interest 
% \ear{this is strange to say,  x in D AND x in R-T;
% instead just say  x in D}
% \ear{
% also  R-T is inconsistent as you called it R-D above instead.
% i assume D refers to the max length of a time series ???
% or, possibly calling that T, so R-T instead everywhere may be better.
% as right now, the curly D and the normal D are looking similar
% and may confuse a reader.}
$\mathbf{x} \in \mathcal{D}$, a time series with $T$ time steps along with a predicted probability distribution $\mathbf{\hat{y}} = f(\mathbf{x})$ over $C$ classes where $\mathbf{\hat{y}} \in [0, 1]^{C}$ and $\sum_{i=1}^{C}\mathbf{\hat{y}}_{i} = 1$. The top predicted class $z$ where
$z = \argmax \mathbf{\hat{y}}$ has predicted confidence $\mathbf{\hat{y}_{z}}$ and is the class for which we seek an explanation.

Our goal is to learn a class-specific saliency map $\boldsymbol{\theta} \in [-1, 1]^T$ for the class of interest $z$ where each element represents the importance of a corresponding time step.
% \ear{HERE YOU USE SYMBOL t for time step..INCONSISTENT}
% time step $t$ with $t$=1, ..., $T$ relative to other classes. 
% A time step is unique if it is important only to the target class $z$ and is not shared with other classes. 
% A time step is important with respect to the probability distribution $P(\mathbf{\hat{y}}_{z}|X)$ for $\mathbf{\hat{y}}_{z} \in \mathcal{Y}$ predicted by $f$. 
Positive values in $\boldsymbol{\theta}$ indicate evidence \textit{for} class $z$, while negative values indicate evidence \textit{against} class $z$.
% $\boldsymbol{\theta}$ reflects the importance of the 
% % \ear{HERE YOU USE SYMBOL $x_t$ for time step..INCONSISTENT.
% % WHILE ONE CAN FIGURE OUT WHAT MAY BE MEANT, IT IS NOT CLEAN. }
% corresponding time step $\mathbf{x}_t$ indicating importance \textit{for} 
% % \ear{what does [0,1] mean here? it should be removed????}
%  class $z$ with a value in interval [0,1] and \textit{against}  class $z$
%  with a value in interval [-1,0].
%  \ear{I am lost be above sentence. it's a real value, or is it in interval? what is that [x,y] integer interval relative to that real value???}
We follow the lead of recent work on \textit{perturbation-based} explanations for time series \cite{Crabbe2021ExplainingTS,Parvatharaju2021LearningSM}, and assume the importance of a time step $t$ should reflect the expected scale of the change of $P(\mathbf{\hat{y}}_{z}|\mathbf{x})$ when $\mathbf{x}$ is perturbed: $|P(\mathbf{\hat{y}}_{z}|\mathbf{x}) - P(\mathbf{\hat{y}}_{z}|\mathbf{\tilde{x}})|$, where $\mathbf{\tilde{x}}$ is perturbed version of $\mathbf{x}$. 
A successfully learned saliency map $\boldsymbol{\theta}$ will 
assign high values to the regions in a time series where perturbations dramatically shift the model's predictions away from $z$.
To ensure that explanations are simple and intuitive, we favor contiguous subsequences within the saliency map, indicating that neighboring time steps likely have similar importance, and that $\boldsymbol{\theta}$ be sparse if possible, assigning no importance to unimportant time steps.

% to the
% % \ear{WHAT DOES SHORT MEAN HERE} 
% short discriminative 
% % \ear{WHAT DOES region MEAN HERE - not defined yet.} 
% sub-sequences, effectively ranking the time steps by their
% % \ear{ DOES IMPACT is equal to the delta  of
% % $|P(\mathbf{\hat{y}}_{z}|\mathbf{x}) - P(\mathbf{\hat{y}}_{z}|\mathbf{\tilde{x}})|$?} 
% impact on $P(\mathbf{\hat{y}}_{z}|\mathbf{x})$. We prefer that saliency map $\boldsymbol{\theta}$ to be sparse, encouraging simpler explanations. 

% \ear{ABOVE: IS IT CLOSE TO ZERO; or is it to have as many time steps
% equal to zero?}

\begin{figure*}[t]
    \centering
    % \hspace{-10mm}
    % \includegraphics[width=\linewidth]{fig/LMUX Architecture_v1.pdf}
    % \includegraphics[width=\linewidth]{fig/LMUX_V1-LMux Rev 2.pdf}
    % \includegraphics[width=\linewidth]{fig/LMUX_Arch.pdf}
    \includegraphics[width=\linewidth]{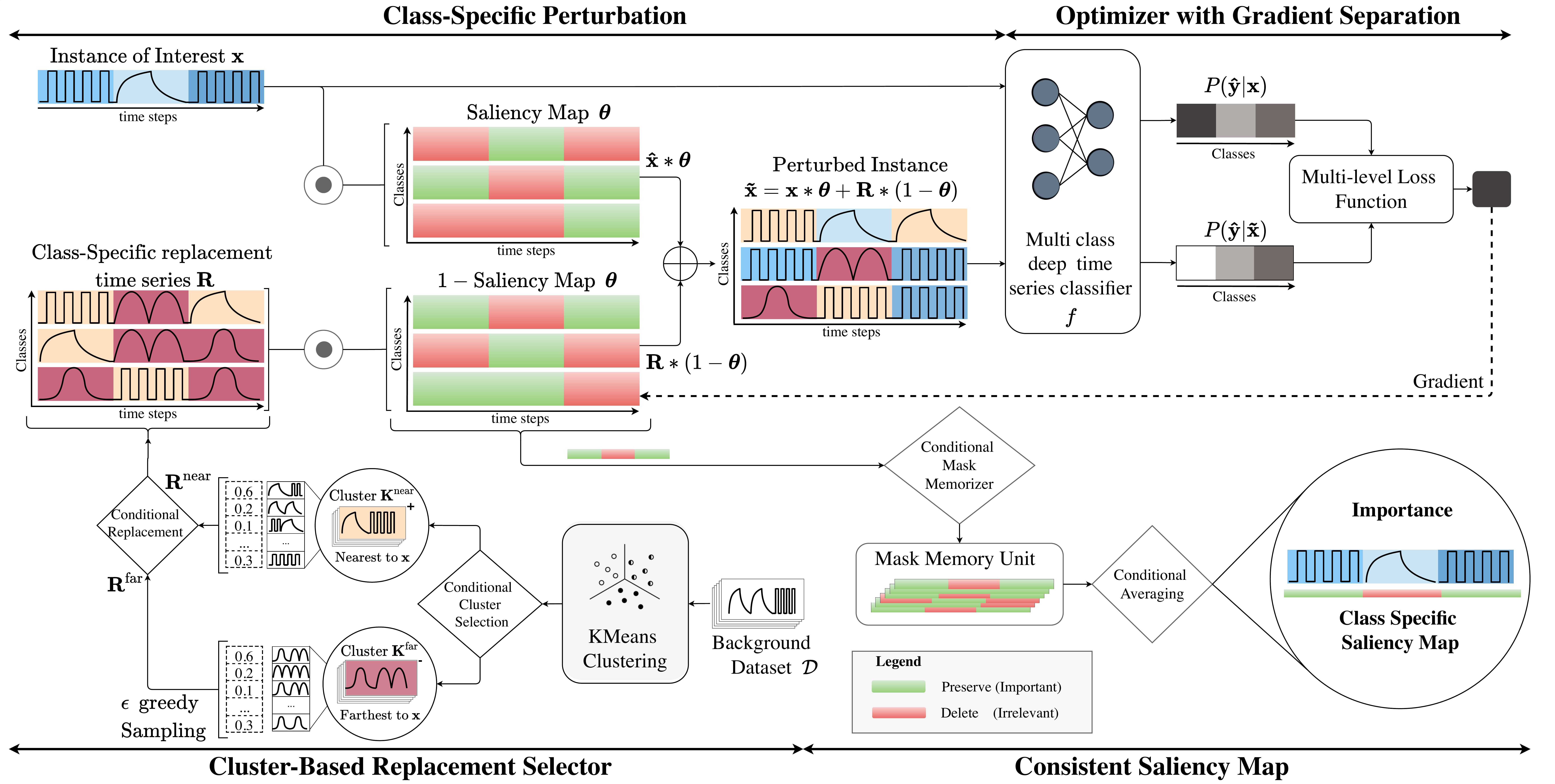}
    % \caption{DEMUX Architecture.}
    \caption{DEMUX Architecture. Three model components collaborate  to produce a saliency map for the predicted class. 1) The Cluster-Based Replacement Selector picks replacement time series. 2) The Class-Specific Perturbation Function uses these replacements to perturb the input. 3) The Mask Memory Unit produces Consistent Saliency Map.}
    %explanations.}
    
    \label{fig:Architecture}
    \vspace{-3mm}
\end{figure*}

\subsection{Proposed Method: DEMUX}
% TODO
% \ear{In figure 2, i cannot find the 3 components?
% where are they. i see one that is a named but where does
% it start or end; the complete left-handside or only the left
% bottom. where are the other two? do use the same font and
% font size to refer to the other 2 components. 
% using the same NAME as you use above:  
%  The Cluster-Based Replacement Selector;
%  The Class-Specific Perturbation Function.
%  Class-Specific Saliency Map }
 
%  \ear{but MAP is an output, and not a component.
%  what is the name of the component that generates the map?
%  is that component called the  ???explanation learner??.
%  if so, add it into the visual. right now
%  the caption and the actual figure are not 100 in synch.}
% RAMESH AND PRATHYUSH --- MAKE THESE EDITS STILL -tom

We propose the Distinct Temporal Multi-class Explainer (DEMUX), the first method to produce class-specific explanations for deep multi-class time series classifiers.
% which derives a class-specific saliency map for the predicted class by a deep multi-class time series classifier $f$.
DEMUX is model agnostic, so can explain any multi-class deep time series classifier $f$'s predictions for an instance $\mathbf{x}$.
By learning to discourage overlapping salient time steps between classes, DEMUX produces saliency maps that highlight the important class-specific time steps.
As illustrated in Figure \ref{fig:Architecture}, DEMUX uses a novel class-specific replacement strategy to perturb time series and explain $f$'s behavior.
DEMUX's learning strategy 
% for producing an effective saliency map $\boldsymbol{\theta}$ is to
encourages $\boldsymbol{\theta}$ to abide by a range of
attractive behaviors and is accomplished via the design of  a novel loss function that leads to explanations that are \textit{unique} to the predicted class, \textit{simple} and \textit{easy} to understand, while being \textit{consistent} between multiple runs. 

DEMUX contains three key components that work together: (1) the \textit{Cluster-Based Replacement Selector} learns which time series from the background dataset $\mathcal{D}$ are the best for replacement-based perturbation.
% The replacements selected are unique to both the instance $\mathbf{x}$ and each class. \tnote{too much detail for here probably}
(2) The \textit{Class-Specific Perturbation Function} learns a saliency map $\boldsymbol{\theta}$ for the top predicted class $z$ relative to other classes by perturbing $\mathbf{x}$ and discovering the impact of each time step on $f$'s prediction probability distribution $P(\mathbf{\hat{y}}|\mathbf{x})$ across all classes. 
(3) A \textit{Mask Memory Unit} to \textit{derive consistent Saliency Maps} which is achieved by building a repository of good saliency maps during learning that matches the predicted probability distribution $P(\mathbf{\hat{y}}|\mathbf{x})$ of $f$.
% We derive $\textit{consistent}$ explanations by performing conditional averaging of saliency maps in the mask memory unit. \tnote{probs don't need so much detail here}
% i.e, the resulting predicted class for $f(\mathbf{\tilde{x})}$ where $\mathbf{\tilde{x}}$ the perturbed instance, matches the predicted class $z$ for $f(\mathbf{\tilde{x}})$  
% \ear{Conditional mean based on what? a little more intuition here
% would be nice.}

\textbf{Cluster-Based Replacement Selector.} \label{sec:cluster_replacement_selector}
Perturbation alters the values of the instance-of-interest $\mathbf{x}$, generating a synthetic time series $\mathbf{\tilde{x}}$. However, choosing which time steps to alter and the scale of the modifications can be highly impactful.
In fact, dramatically changing $\mathbf{x}$ often creates an out-of-distribution perturbation, on which $f$'s predictions are untrustworthy. PERT \cite{Parvatharaju2021LearningSM} overcomes this challenge by creating in-distribution perturbations via replacement time series $\mathbf{r}$, which replace time steps in $\mathbf{x}$ with time steps from background instances in $\mathcal{D}$.
Inspired by PERT, we select a pair of time series from the background dataset $\mathcal{D}$ to provide evidence both $\textit{for}$ and $\textit{against}$ the class of interest $z$. However, multi-class replacement time series sampling is non-trivial as there can be many classes to choose from.
% Sampling from the valid set of classes to derive both $\textit{for}$ and $\textit{against}$  evidences for class of interest $C$ is crucial \tnote{I don't know what it means to sample from a set of classes? And what makes a set of classes "valid?"}. 
% Following PERT's general strategy, the instance of interest $X$ will be explained by learning the degree to which $X$ is interpolated with each chosen time series.
% produced a new method to perturb the values of a time series for univariate time series, but it is not suited well for multi-class time series context. DEMUX extends this method to consider class-specific perturbations.
% Thus, the first component of DEMUX chooses which time series is appropriate to use from the background dataset for replacement of the time steps of instance $X$.

The best replacement time series may also vary by time step and per class, so we select replacement series $\mathbf{r}$ on a per-class and time step-by-time step basis.
To provide evidence both \textit{for} and \textit{against} the class of interest $z$, the \textit{Cluster-Based Replacement Selector} employs a clustering on $\mathcal{D}$---we use \textit{KMeans} with \textit{Kneedle} \cite{Satopaa2011FindingA} to find a suitable number of clusters $k$ for the given dataset $\mathcal{D}$ in our experiments.
% constructs $P$ clusters using KMeans. We find an appropriate number of clusters $K$ using the Kneedle algorithm . 
%
%From these $k$ clusters, for each class, the Cluster-Based Replacement Selector chooses two clusters of time series representatives $\mathbf{K^\text{near}}$ and $\mathbf{K^\text{far}}$. 
%
Thereafter, for each class, 
the Cluster-Based Replacement Selector chooses two clusters of time series representatives $\mathbf{K}^\text{near}$ and $\mathbf{K}^\text{far}$ from  these $k$ clusters. 
%
% with $j_\text{near}$ and $j_\text{far}$ as respective centroids. 
$\mathbf{K}^\text{near}$ denotes a cluster nearest to $\mathbf{x}$ and $\mathbf{K}^\text{far}$ a cluster furthest from $\mathbf{x}$ based on the \textit{Euclidean} distance between $\mathbf{x}$ and the centroid of the two clusters, respectively.
Then, we pick one replacement time series from $\mathbf{K^\text{near}}$ and one from $\mathbf{K}^\text{far}$
\begin{align} 
% \hspace{-10mm}
  \label{eq:Greedy}
    \mathbf{r}^\text{far} :=
    \begin{cases}
        \mathbf{r}^\text{far}\ & \text{with probability } 1-\epsilon\\ 
        \text{random } \mathbf{r} \in \mathbf{K}^\text{far} & \text{with probability } \epsilon
    \end{cases}
\end{align} 
where $\mathbf{r}^\text{far}$ is one time series sampled from $\mathbf{K^\text{far}}$ with probability $\epsilon$. 
% \tnote{this is confusing still, if there's some probability r far remains unchanged, how is it initialized?}
% \in \mathbb{R}^{C \times T}$ is a collection of $C$ replacement time series (one per class)
% \tnote{would we lose anything if we just call it $R^\text{far}$? the $c{\times}T$ stuff is confusing and doesn't seem necessary to me.. and usually if it's going to describe the shape, you just say $R^\text{far} \in \mathbb{R}^{c\times T}$ is a collection of one replacement time series per class $c$ } defined in equation  
% sampled from $\mathbf{K^\text{far}}$.
We also sample $\mathbf{r}^\text{near}$ from the nearest cluster $\mathbf{K^\text{near}}$. 
We repeat this process for each of the $C$ classes, accumulating the replacement series into respective matrices $\mathbf{R}^{near}, \mathbf{R}^{far} \in \mathbb{R}^{C \times T}$, denoting two replacement time series $\mathbf{r}^{near}$ and $\mathbf{r}^{far}$ per class.
% \tnote{this paragraph comes out of nowhere... you need to reference the $R$ in the text as something you're trying to compute up above: Our first goal is to select two time series per class $c$. One time series, $R_c$, will help find evidence for class $c$. The other, $R_n$ will help... }. 
We employ epsilon greedy-based prioritized sampling \cite{Schaul2016PrioritizedER} to explore and learn which replacement time series sampled from $\mathcal{D}$ are best for instance-of-interest $\mathbf{x}$

\textbf{Class-Specific Perturbation.}\label{sec:perturbation_function}
Next, we derive a class-specific saliency map $\boldsymbol{\theta}$ for $z$, the top predicted class, relative to other classes in $\mathbf{\hat{y}}$.
We perturb the instance of interest $\mathbf{x}$ by performing time step and class-specific replacement using replacement time series matrices and a saliency map $\boldsymbol{\theta}$. Equation \ref{eq:Matrix_Pert} shows a class-specific perturbation where $\mathbf{x}^*_i$ is the perturbed instance for the $i$-th class. 
% we perturb $\tilde{X_{c,n}}$ \tnote{put tildes over just X, not the whole thing! $\tilde{X}_{c, n}$}, which is a matrix of modified versions of instance $X$ using our \textit{Class-Specific Perturbation Function} and per-class Replacement time series from clusters $R_N$ and $R_F$.
${\mathbf{x}^*_i}$ is passed to classifier $f$ to observe the effects of the perturbation on $f$'s predicted distribution $P(\mathbf{\hat{y}}|\mathbf{x})$.

We \textit{learn} a perturbation function $\Phi: \mathbb{R}^{C \times T} \to \mathbb{R}^{C \times T}$ where the key component of $\Phi$ is a two-dimensional parameterized matrix 
% $\boldsymbol{\theta} \in \mathbb{R}^{C{\times}T} [-1, 1]^T$, where $C$ is the number of classes and $T$ is the number of time steps per class. 
$\boldsymbol{\theta} \in [-1, 1]^{C \times T}$. An element $\theta_{i,t}$ of this parameterized matrix represents the importance value of time step $t$ for class
% \ear{Earlier classes were denoted by z, and not little x? INCONSISTENT}
$i$ relative to other classes. 
% $\theta$ serves as the class-specific $|\mathcal{Y}|$ saliency maps with $t$ time steps each. 
Values in $\boldsymbol{\theta}$ close to 1 indicate strong evidence \textit{for} the respective class $i$. Values close to -1 indicate evidence \textit{against} the class $i$. Values near 0 imply no importance.

To consider both evidence \textit{for} and \textit{against} each class, our perturbation function $\Phi$ adaptively learns to replace values at each time step with replacement time series matrices. We thus  replace the time step with $\mathbf{R^\text{near}}$, the representative of time series from the cluster nearest to $\mathbf{x}$ for $\theta_{i,t} < 0$. 
% \tnote{I'm lost.. you combine them into $R$ above.. now there's a new thing: $R^\text{near}$??}
Similarly, when $\theta_{i,t} \ge 0$, $\mathbf{x}$ is replaced with the corresponding time step of $\mathbf{R^\text{far}}$, the representative of time series from the cluster farthest to $\mathbf{x}$. This way, DEMUX learns the degree of sensitivity of each time step per class.
The function $\Phi$ generates perturbation $\mathbf{\tilde{x}}$ by performing time step-specific interpolation for each class:
% \tiny
% \small
% \footnotesize	
\begin{equation}\label{eq:Matrix_Pert}
    \resizebox{0.91\hsize}{!}{%
% \resizebox{\linewidth}{!}{
% \label{eq:Matrix_eq}
    $\mathbf{\tilde{x}} = \boldsymbol{\theta}_{i} \odot \mathbf{x} + (1 - \boldsymbol{\theta}_{i}) \odot (\mathbbm{1}_{\boldsymbol{\theta}_{i} < 0} \odot \mathbf{R}^\text{near}_{i} + \mathbbm{1}_{\boldsymbol{\theta}_{i} \geq 0} \odot \mathbf{R}^\text{far}_{i}) + g$
% \begin{pmatrix}
% \theta_{1,n} \odot X + (1 - \theta_{1,n} \odot (\mathbbm{1}_{\theta_{1,n} < 0} \odot R_{N{(1,n)}} + \mathbbm{1}_{\theta_{1,n} \geq 0} \odot R_{F(1,n)}) + g_1 \\
% \theta_{2,n} \odot X + (1 - \theta_{2,n} \odot (\mathbbm{1}_{\theta_{2,n} < 0} \odot R_{N{(2,n)}} + \mathbbm{1}_{\theta_{2,n} \geq 0} \odot R_{F(2,n)}) + g_2 \\
% \theta_{3,n} \odot X + (1 - \theta_{3,n} \odot (\mathbbm{1}_{\theta_{3,n} < 0} \odot R_{N{(3,n)}} + \mathbbm{1}_{\theta_{3,n} \geq 0} \odot R_{F(3,n)}) + g_3 \\
% \vdots\\
% % \theta_(m-1) \odot X + (1 - \theta_(m-1) \odot (\mathbbm{1}_{\theta_(m-1) < 0} \odot R_{m-1} + \mathbbm{1}_{\theta_(m-1) \geq 0} \odot R_{O(m-1)}) + g_{m-1} \\
% \theta_{C,n} \odot X + (1 - \theta_{C,n} \odot (\mathbbm{1}_{\theta_{C,n} < 0} \odot R_{N{(C,n)}} + \mathbbm{1}_{\theta_{C,n} \geq 0} \odot R_{F(C,n)}) + g_C \\
% \end{pmatrix}
    }
\end{equation} 

% \tnote{again, the $c\times n$ thing is too confusing, you could just index instead: "$\tilde{X}_i$ is the perturbation for the $i$-th class." Then describe everything just for one class! $R^\text{near}_i$ is for the $i$-th class too! Please make this fix. You can say one time, that $\theta \in \mathbb{R}^{C \times T}$, assuming $C$ is the number of classes and $T$ is the number of time steps.}

% \begin{equation*}\label{eq:Matrix_Pert}
% $\tilde{X}_{C,n} = $
% \begin{pmatrix}

% \theta_{1,n} \odot X + (1 - \theta_{1,n} \odot (\mathbbm{1}_{\theta_{1,n} < 0} \odot R_{N{(1,n)}} + \mathbbm{1}_{\theta_{1,n} \geq 0} \odot R_{F(1,n)}) + g_1 \\

% \theta_{2,n} \odot X + (1 - \theta_{2,n} \odot (\mathbbm{1}_{\theta_{2,n} < 0} \odot R_{N{(2,n)}} + \mathbbm{1}_{\theta_{2,n} \geq 0} \odot R_{F(2,n)}) + g_2 \\

% \theta_{3,n} \odot X + (1 - \theta_{3,n} \odot (\mathbbm{1}_{\theta_{3,n} < 0} \odot R_{N{(3,n)}} + \mathbbm{1}_{\theta_{3,n} \geq 0} \odot R_{F(3,n)}) + g_3 \\

% \vdots\\

% % \theta_(m-1) \odot X + (1 - \theta_(m-1) \odot (\mathbbm{1}_{\theta_(m-1) < 0} \odot R_{m-1} + \mathbbm{1}_{\theta_(m-1) \geq 0} \odot R_{O(m-1)}) + g_{m-1} \\

% \theta_{C,n} \odot X + (1 - \theta_{C,n} \odot (\mathbbm{1}_{\theta_{C,n} < 0} \odot R_{N{(C,n)}} + \mathbbm{1}_{\theta_{C,n} \geq 0} \odot R_{F(C,n)}) + g_C \\

% \end{pmatrix}
% }
% \end{equation*} 

% \normalsize

Here, $\mathbbm{1}_{\boldsymbol{\theta}_{i} < 0}$ is a matrix-wise indicator function that returns 1 for elements of $\boldsymbol{\theta}_{i}$ that are less than 0 and $\mathbbm{1}_{\boldsymbol{\theta}_{i} \geq 0}$ returns 1 for elements of $\boldsymbol{\theta}_{i}$ greater than or equal to 0. $\odot$ is the Hadamard product. For readability, we refer to this operation as $\Phi(\mathbf{x}; \boldsymbol{\theta})$. 
Inspired by \cite{Fong2017InterpretableEO},
We add a small amount of Gaussian noise $g$ to avoid overfitting $\boldsymbol{\theta}$ to extremely specific values.
Using Equation \ref{eq:Matrix_Pert}, the final values of $\mathbf{\tilde{x}}$, the perturbed version of $\mathbf{x}$, are thus interpolations between the original time steps of $\mathbf{x}$ and the replacement series $\mathbf{R}^\text{near}_{i}$ or $\mathbf{R}^\text{far}_{i}$  where $i$ is the $i$-th class in $\mathbf{\hat{y}}$ according to the scale of the corresponding value in $\boldsymbol{\theta}_{i}$. The conditional interval specific operations on $\boldsymbol{\theta}$ are necessary to derive evidence both \textit{for} and \textit{against} class $z$.
% \tnote{this paragraph doesn't include much intuition about WHY we do things this way.. the replacement with near vs. far is described effectively, but ideally we should motivate this design CONCEPTUALLY, too}

% We initialize the values of $\theta$ uniformly between $-1e^{-2}$ and $1e^{-2}$ to encode no prior assumptions on which time steps are most important and steadily learn to insert only the most crucial time steps to the prediction $P(\hat{y}|X)$. This way, we encourage the system to provide simpler explanations, as the default sum of $\theta$ is small.

% \begin{equation}
%     \label{eq:perturb}
%     \sum_{c=0}^{M} \tilde{X_c} = \theta_c \odot X + (1 - \theta_c) \odot (\mathbbm{1}_{\theta < 0} \odot R_C + \mathbbm{1}_{\theta \geq 0} \odot R_O) + g
% \end{equation}

% \begin{equation}
%     L_\text{Probability Distribution(PD)} = \lambda_{1} \left( \sum_{c=0}^{M} \left( \frac{1}{\left \|\hat{X}  \right \|} \sum_{i=0}^{N}(f_{c}(\hat{X})-f_{c}(f_{p}(\hat{X};\theta_c)))^2 \right ) \right )
% \end{equation}

% $\tilde{X}_{c, n}$
\textbf{Learning Class-Specific Explanation.}
\label{sec:optimizing}
Class-specific saliency values $\boldsymbol{\theta} \in \mathbb{R}^{C{\times}T}$ are learned using a novel loss function (Equation \ref{eq:total_loss}) containing five key components.
% $L_{Prev}$, $L_{Max}$, $L_{Budget}$, $L_{TReg}$,  and $L_{SSD}$.
We 
% jointly \ear{joint with what? is it C times the same function , or what is joint with what?} 
optimize the loss function to derive a simple instance-specific explanation 
% that is , simple, meaningful, consistent and unique explanation 
for the predicted distribution $P(\mathbf{\hat{y}}|\mathbf{x})$ for $\mathbf{x}$. % by a multi-class deep time series classifier $f$. 

First, we encourage
\textit{Class-Specificity}: Overlapping salient regions between classes should be minimized. We design an objective called \textit{shared saliency deletion} $L_\text{SSD}$,
which learns to remove shared saliency amongst the top predicted class $z$ and the rest of the classes, proportional to the respective class probability. This uniquely differentiates the top class salient time steps from the other classes. 
% This way, we uniquely differentiate the top class's important values from those of the other classes.
DEMUX learns unique class-specific saliency map $\boldsymbol{\theta}_{z}$ for the top predicted class $z$ by altering $\boldsymbol{\theta}_{z}$ with respect to the corresponding importance values at the same time steps from other classes. Consider $\theta_{z,t}$ where $t$ is the time step, if the importance value for time step $t$ is high for multiple classes, it is considered to be a shared salient time step. The importance value $\theta_{z,t}$ is reduced by the importance value of time step $t$ from $\boldsymbol{\theta}_{i, t}$ where $i \neq z$ represents a different class, and is weighted by the respective class confidence $P(\mathbf{\hat{y}}_{i}|\mathbf{x})$ to counter the influence of least confidence classes. Intuitively, $L_\text{SSD}$ improves the importance of unique salient time steps and reduces the importance of shared salient time steps. 
% \tnote{this last sentence---no need to repeat above info, but you do need to explain what it does... or if you don't think there's more to explain after the last couple sentences, just delete this sentence. This is such an important paragraph that is way too hidden}

\begin{equation}
\label{eqn:ssd}
    L_{SSD} = \lambda_{5} \left ( \frac{1}{\left \|{\mathcal{Y}} \right \|} \left( \boldsymbol{\theta}_{z} - \left \| \sum_{\substack{\\ i\neq z}}^{\mathcal{Y}} (\boldsymbol{\theta}_{z} - P(\mathbf{\hat{y}}_{i}|\mathbf{x}) * \boldsymbol{\theta}_i) \right \| \right)^2 \right )
\end{equation}

Second, to derive instance-specific explanations from $f$, it is crucial to preserve \textit{Local Fidelity} by preserving $P(\mathbf{\hat{y}}|\mathbf{x})$ for $\mathbf{x}$. Any changes to the predicted probability distribution $P(\mathbf{\hat{y}}|\mathbf{x})$ creates a disconnect between the derived explanation $\boldsymbol{\theta}_{z}$ and instance-of-interest $\mathbf{x}$.
% top predicted confidence fails to guarantee 
% \tnote{you have to say why}.
The second component of our loss function, $L_{Prev}$, encourages the perturbation function $\Phi$ to produce perturbations for the top class $z$ while preserving the distribution $P(\mathbf{\hat{y}}|\mathbf{x})$ for $\mathbf{x}$.
To achieve this preservation, we use the Kullback-Liebler (KL) Divergence (Equation \ref{eq:KL_LOSS}) to measure how the predicted distribution for the perturbed instance $P(\mathbf{\hat{y}}|\mathbf{\tilde{x}})$ differs from the reference distribution $P(\mathbf{\hat{y}}|\mathbf{x})$. 
\begin{equation}
\label{eq:KL_LOSS}
    L_\text{Prev} = \lambda_{1} \left(  \left(  f(\mathbf{x}) - f(\Phi(\mathbf{x};\boldsymbol{\theta}_{z})) \right )_{KL} \right )
\end{equation}
% TODO
% \ear{ARE THE 4 technical challenges listed in the intro
% ever mentioned anywhere in your method section????
% the low prob may be one - if so, connect it to intro .}

Third, to generate preserving low probabilities is a challenge as there are many possible perturbations that result in a low probabilities. Thus, for classes other than highest confidence class $z$ we encourage the saliency map $\boldsymbol{\theta}$ to highlight the time steps that are responsible for maximizing the specific-class probability represented by Equation \ref{eq:L_MAX}.
Intuitively, we \textit{preserve} the confidence for the top predicted class $z$, to derive explanations specific to instance-of-interest $\mathbf{x}$. For the rest of the classes, we maximize the confidences to 
% instead play a \textit{maximization game} \tnote{it's weird to say we "play a game" I think---let's just say what we do. REWRITE THIS}
find the minimal salient time steps that causes the prediction probability for the respective class to increase significantly. By making use of $\textit{gradient separation}$ and $\textit{multi-objective loss}$,  we 
learn saliency maps for $z$ and the other classes distinctively. 
% \tnote{I have no idea what these 2 sentences mean}
% \ear{Above, what is mulit-level - you mean first one loss, and based on it to state the next loss- point more directly what you mean when you throw around terms...}

\begin{equation}
\label{eq:L_MAX}
    L_\text{Max} = \lambda_{2} \left( \frac{1}{\left \|{\mathcal{Y}} \right \|} \sum_{i\neq z}^{\mathcal{Y}} \left(1 - f(\Phi(\mathbf{x};\boldsymbol{\theta}_{i})) \right ) \right )
\end{equation}

Fourth, to encourage 
% \ear{What does
% legible mean? did you mean meaningful? or, maybe the term "simple" would
% be better here?}
% legible explanations with 
% \ear{Why smaller regions?}
simple explanations, with minimal salient time steps per class, we incorporate the $L_{Budget}$ loss  per class, 
similar to PERT \cite{Parvatharaju2021LearningSM} and DYNAMASK \cite{Crabbe2021ExplainingTS} methods.
This component encourages the values of $\boldsymbol{\theta}$ to be as small as possible; intuitively, values that are not important should be close to 0 per our problem definition.

% \ear{Font in equation looks bigger than regular text; but is not a critical
% format thing to fix right now. getting prose right is much more important.}

% \ear{Below is it really  c=0 to y?}
\begin{equation}
\label{eq:budget}
L_\text{Budget} = \lambda_{3} \left ( \frac{1}{\left \|{\mathcal{Y}} \right \|} \sum_{i=0}^{\mathcal{Y}} \left ( \frac{1}{\left \| T \right \|}\sum_{t=0}^{T}\left | \theta_{i,t} \right | \right)  \right)
\end{equation}

Fifth, we encourage the saliency map to be \textit{Temporally Coherent}: neighboring time steps should generally have similar important. To achieve this coherence, we add a time series regularizer $L_\text{TReg}$ per class that minimizes the squared difference between neighboring saliency values:
\begin{equation}
\label{eq:tv_norm}
L_\text{TReg} = \lambda_{4} \left ( \frac{1}{\left \|{\mathcal{Y}} \right \|} \sum_{i=0}^{\mathcal{Y}} \left ( \frac{1}{\left \| T \right \|} \sum_{t=0}^{T-1} (\theta_{i,t}-\theta_{i,t+1})^{2} \right) \right)
\end{equation}

% \subsubsection{Class-Specific Evidence.}
% \tnote{weird to have a new subsection here.. we were working towards a loss function, then this takes a break!}

% In the example figure \ref{fig:Venn}, Target Class A saliency is shared with the classes B and C as highlighted in the green color \tnote{I am not a fan of the venn diagram, it doesn't help us understand "overlap" better.. that's just the picture everyone thinks of for overlap. If you want a picture for this, it should be a time series with a saliency map}.

% \begin{equation}
% \label{eq:L_MAX}
%     L_\text{Max} = \lambda_{2} \left( \frac{1}{\left \|{C} \right \|} \sum_{c\neq \hat{y}} \left(1 - f(f_{p}(\hat{X};\theta_c)) \right ) \right )
% \end{equation}
% \tnote{trying a clearer notation for the sum}

%  \begin{figure}[t]
%   \centering
%   \hspace*{-0.9cm}  
%   \includegraphics[scale=0.5]{fig/VennDiagram.pdf} 
%     %  \vspace{-5mm}
%       \caption{Learning a $Class-Specific$ Saliency for the Target Class A relative to other classes.}
%     \label{fig:Venn}
%  \end{figure}

% $\mathbb{ \lambda_{1},\lambda_{2},\lambda_{3},\lambda_{4},\lambda_{5}}$ are hyperparameters for each of these components.

Finally, all the loss terms are summed, each being scaled by a $\lambda$ coefficient to balance the components depending on task-specific preferences in explanation behavior.
% We scale the final loss function according to the sampling weights $w_F$ and $w_N$. 
Minimizing the total loss in Equation 
\ref{eq:total_loss} leads to simple, class-specific saliency maps.

% \footnotesize
\begin{equation}
   \label{eq:total_loss}
   L(P(\mathbf{\hat{y}}|\mathbf{x}); \boldsymbol{\theta}) = L_\text{SSD} + L_\text{Prev} + L_\text{Max} + L_\text{Budget} + L_\text{TReg}
   \end{equation} 
%   \tnote{The P() thing wasn't used above, so not clear here.}
% \normalsize
% \subsubsection{Ensemble of Saliency Maps.}

\textbf{Mask Memory Unit for Consistent Saliency Maps.}
A common problem with perturbation-based methods \cite{Parvatharaju2021LearningSM, Crabbe2021ExplainingTS, Fong2017InterpretableEO} is high variance between explanations, even given the same inputs multiple times. Also, during learning between epochs, the saliency maps can change dramatically \cite{Kumar2020ProblemsWS}. This makes training  challenging, in particular, 
to know when to stop training. We address this problem  using a \textit{Mask Memory Unit}, in which we store ``good'' saliency maps throughout training. A saliency map is considered to be good if the resulting prediction $\argmax f(\Phi(\mathbf{\tilde{x}}))$ for the perturbed instance
% \ear{WHAT does i=z notation mean here? it is new.} 
$\mathbf{\tilde{x}}$ is identical to the predicted class $z$ for the instance of interest  $\mathbf{x}$. 
% For each epoch when the perturbed instance generates prediction probability distribution $f(f_{p}(\hat{X};\theta_c))$ is like the original opaque box model's prediction $f(\hat{X})$ then we classify it has good saliency map and we added it to the repository \tnote{sentence way too long... makes no sense. break it down and be specific... how do you know when distributions are similar, how do you classify it? how do you define "good"?}. 
The final saliency map generated corresponds to the average of the last few (10 or more) saliency maps in the repository. This significantly reduces the variance between the saliency maps offered by our model as it results in generating consistent explanations for same time series instance across multiple runs.

\begin{table}[h]
\centering
\resizebox{\linewidth}{!}{
\begin{tabular}{@{}lccccccccc@{}}
\toprule
\textbf{Dataset}          & \textsc{ACSF1} & \textsc{Plane} & \textsc{Trace} & \textsc{ECG5000} & \textsc{Meat} \\ \midrule
Num. Train Instances  & 100  & 105       & 100       & 500         & 60  \\
Num. Test Instances   & 100  & 105       & 100       & 4500        & 60  \\
Num. Time steps        & 1460 & 144       & 275       & 140         & 500 \\
Num. Classes          & 10   & 7         & 4         & 5           & 3   \\
\bottomrule
\end{tabular}
}

\caption{Dataset summary statistics.} 
\label{tab:mc_dataset}
\vspace{-5mm}
\end{table}

\begin{table*}[hbt!]
\centering
\begin{tabular}{@{}lllllllllll@{}}
\toprule
\multicolumn{1}{l}{\multirow{3.5}{*}{\textbf{Methods}}} & \multicolumn{10}{c}{\textbf{Datasets}} \\ \cmidrule(l){2-11} 
\multicolumn{1}{c}{} & \multicolumn{2}{c|}{\textsc{ECG5000}} & \multicolumn{2}{c|}{\textsc{Plane}} & \multicolumn{2}{c|}{\textsc{Trace}} & \multicolumn{2}{c|}{\textsc{ACSF1}} & \multicolumn{2}{c}{\textsc{Meat}} \\ \cmidrule(l){2-11} 
\multicolumn{1}{c}{} & \multicolumn{1}{c}{\textbf{AUC $\uparrow$}} & \multicolumn{1}{c|}{\textbf{IoU $\downarrow$}} & \multicolumn{1}{c}{\textbf{AUC $\uparrow$}} & \multicolumn{1}{c|}{\textbf{IoU $\downarrow$}} & \multicolumn{1}{c}{\textbf{AUC $\uparrow$}} & \multicolumn{1}{c|}{\textbf{IoU $\downarrow$}} & \multicolumn{1}{c}{\textbf{AUC $\uparrow$}} & \multicolumn{1}{c|}{\textbf{IoU $\downarrow$}} & \multicolumn{1}{c}{\textbf{AUC $\uparrow$}} & \multicolumn{1}{c}{\textbf{IoU $\downarrow$}} \\ \midrule
RISE $\cite{Petsiuk2018RISERI}$ & 0.60 (.005) & \multicolumn{1}{l|}{15.4\%} & 0.25 (.009) & \multicolumn{1}{l|}{7.52\%} & 0.49 (.004) & \multicolumn{1}{l|}{19.8\%} & 0.16 (.007) & \multicolumn{1}{l|}{19.9\%} & 0.17 (.010) & 29.3\% \\
LEFTIST $\cite{Guillem2019AgnosticLE}$ & \textbf{0.84 (.003)} & \multicolumn{1}{l|}{\textbf{22.5\%}} & 0.39 (.011) & \multicolumn{1}{l|}{33.8\%} & 0.31 (.007) & \multicolumn{1}{l|}{36.9\%} & 0.17 (.014) & \multicolumn{1}{l|}{36.6\%} & 0.36 (.006) & 30.3\% \\ \midrule
LIME $\cite{Ribeiro2016WhySI}$ & 0.78 (.012) & \multicolumn{1}{l|}{29.9\%} & 0.50 (.005) & \multicolumn{1}{l|}{30.8\%} & 0.46 (.006) & \multicolumn{1}{l|}{37.1\%} & 0.19 (.006) & \multicolumn{1}{l|}{37.9\%} & 0.15 (.002) & 38.4\% 
\\
TSMULE $\cite{Schlegel2021TSMULELI}$ & 0.43 (.017) & \multicolumn{1}{l|}{11.2\%} & 0.27 (.009) & \multicolumn{1}{l|}{7.39\%} & 0.38 (.015) & \multicolumn{1}{l|}{19.5\%} & 0.05 (.008) & \multicolumn{1}{l|}{13.1\%} & 0.12 (.003) & 31.7\% \\
SHAP $\cite{NIPS2017_7062}$ & 0.45 (.133) & \multicolumn{1}{l|}{12.7\%} & 0.38 (.011) & \multicolumn{1}{l|}{20.2\%} & 0.52 (.045) & \multicolumn{1}{l|}{30.6\%} & 0.04 (.005) & \multicolumn{1}{l|}{18.7\%} & 0.25 (.002) & 24.9\% \\
TIMESHAP $\cite{bento2020timeshap}$ & 0.73 (.175) & \multicolumn{1}{l|}{20.5\%} & 0.27 (.023) & \multicolumn{1}{l|}{9.6\%} & 0.31 (.067) & \multicolumn{1}{l|}{17.6\%} & 0.02 (.089) & \multicolumn{1}{l|}{17.2\%} & 0.15 (.033) & 19.2\%
\\\midrule
MP  $\cite{Fong2017InterpretableEO}$ & -0.38 (.021) & \multicolumn{1}{l|}{93.8\%} & 0.16 (.001) & \multicolumn{1}{l|}{66.6\%} & 0.34 (.009) & \multicolumn{1}{l|}{13.4\%} & 0.06 (.011) & \multicolumn{1}{l|}{12.7\%} & 0.23 (.044) & 2.25\% \\
DYNAMASK $\cite{Crabbe2021ExplainingTS}$ & 0.69 (.005) & \multicolumn{1}{l|}{7.41\%} & 0.25 (.008) & \multicolumn{1}{l|}{{3.23\%}} & 0.23 (.007) & \multicolumn{1}{l|}{4.54\%} & 0.13 (.007) & \multicolumn{1}{l|}{{1.62\%}} & 0.05 (.001) & {0.27\%} \\
PERT $\cite{Parvatharaju2021LearningSM}$ & 0.69 (.010) & \multicolumn{1}{l|}{32.8\%} & 0.13 (.007) & \multicolumn{1}{l|}{21.2\%} & 0.18 (.001) & \multicolumn{1}{l|}{40.4\%} & 0.16 (.047) & \multicolumn{1}{l|}{37.9\%} & 0.33 (.021) & 21.1\% \\
DEMUX & 0.78 (.007) & \multicolumn{1}{l|}{{0.24\%}} & \textbf{0.71 (.001)} & \multicolumn{1}{l|}{\textbf{4.31\%}} & \textbf{0.52 (.004)} & \multicolumn{1}{l|}{\textbf{2.44\%}} & \textbf{0.45 (.012)} & \multicolumn{1}{l|}{\textbf{2.11\%}} & \textbf{0.58 (.007)} & \textbf{2.46\%}\\
\bottomrule
\end{tabular}
\caption{Performance of the AUC-Difference $\uparrow$ and IoU $\downarrow$ metrics with the FCN model. Parentheses indicate $\sigma$.}
\label{tab:fcn_mux_auc_table}
\end{table*}

% https://wandb.ai/xai/ICDM/table?workspace=user-rameshdoddaiah

% \begin{table*}[hbt!]
% {
% \centering
% \begin{tabular}{lccccc}
% \toprule
% \multirow{2.6}{*}{\textbf{Methods}} & \multicolumn{5}{c}{\textbf{Datasets}}\\
% \cmidrule{2-6 }
%  &\textsc{ECG5000} & \textsc{Plane} & \textsc{Trace} & \textsc{ACSF} & \textsc{Meat}\\

% \midrule

% RISE \cite{Petsiuk2018RISERI}  & \ 15.4\%,  & 7.5\% & 19.8\% & 19.9\% & 29.3\% \\
% \midrule
% LIME \cite{Ribeiro2016WhySI}   & \29.9\% & \ 30.8\% & 37.1\% & 37.9\% & 38.4\% \\
% \midrule
% SHAP \cite{NIPS2017_7062}  & 12.7\% & 20.2\% & 30.6\% & 18.7\% & 24.9\%\\
% \midrule
% MP  \cite{Fong2017InterpretableEO}  &93.8\% & 66.6\% & 13.4\% &12.7\% & 2.2\% \\
% \midrule
% LEFTIST \cite{Guillem2019AgnosticLE} & \22.5\% & 33.8\% & 36.9\% & 36.6\% & 30.3\% \\
% \midrule
% Dynamask \cite{Crabbe2021ExplainingTS} &7.4\% & 3.2\% &4.5\% & 1.6\% &0.27\%\\
% \midrule
% PERT \cite{Parvatharaju2021LearningSM} & 32.8\%  &21.2\% &40.4\% & 37.9\% &21.1\% \\
% \midrule
% DEMUX  &\textbf{0.24\%} & \textbf{4.3\%} & \textbf{2.4\%} &\textbf{2.1\%} &\textbf{2.4\%}\\
% \bottomrule
% \end{tabular}
% }
% \caption{Performance of the IoU $\downarrow$ metric with the FCN black-box model. Parentheses indicate $\sigma$.} 
% \label{tab:mux_auc_table}
% \vspace{-7mm}
% \end{table*}

% \begin{figure*}
%     \centering
%     \includegraphics[scale=0.23]{fig/AUC.png}
%     % \caption{DEMUX Architecture}
%     \label{fig:Architecture}
% \end{figure*}

% \label{tab:auc}

\begin{table*}[hbt!]
\centering
\begin{tabular}{@{}lllllllllll@{}}
\toprule
\multicolumn{1}{l}{\multirow{3.5}{*}{\textbf{Methods}}} & \multicolumn{10}{c}{\textbf{Datasets}} \\ \cmidrule(l){2-11} 
\multicolumn{1}{c}{} & \multicolumn{2}{c|}{\textsc{ECG5000}} & \multicolumn{2}{c|}{\textsc{Plane}} & \multicolumn{2}{c|}{\textsc{Trace}} & \multicolumn{2}{c|}{\textsc{ACSF1}} & \multicolumn{2}{c}{\textsc{Meat}} \\ \cmidrule(l){2-11} 
\multicolumn{1}{c}{} & \multicolumn{1}{c}{\textbf{AUC $\uparrow$}} & \multicolumn{1}{c|}{\textbf{IoU $\downarrow$}} & \multicolumn{1}{c}{\textbf{AUC $\uparrow$}} & \multicolumn{1}{c|}{\textbf{IoU $\downarrow$}} & \multicolumn{1}{c}{\textbf{AUC $\uparrow$}} & \multicolumn{1}{c|}{\textbf{IoU $\downarrow$}} & \multicolumn{1}{c}{\textbf{AUC $\uparrow$}} & \multicolumn{1}{c|}{\textbf{IoU $\downarrow$}} & \multicolumn{1}{c}{\textbf{AUC $\uparrow$}} & \multicolumn{1}{c}{\textbf{IoU $\downarrow$}} \\ \midrule
RISE $\cite{Petsiuk2018RISERI}$ & -0.34 (.005) & \multicolumn{1}{l|}{3.71\%} & 0.31 (.009) & \multicolumn{1}{l|}{19.8\%} & 0.25 (.004) & \multicolumn{1}{l|}{1.73\%} & 0.14 (.005) & \multicolumn{1}{l|}{20.5\%} & 0.45 (.010) & 1.42\% \\
 LEFTIST $\cite{Guillem2019AgnosticLE}$ & 0.80 (.002) & \multicolumn{1}{l|}{29.9\%} & 0.36 (.005) & \multicolumn{1}{l|}{28.8\%} & 0.91 (.006) & \multicolumn{1}{l|}{2.16\%} & 0.14 (.019) & \multicolumn{1}{l|}{27.3\%} & 0.72 (.002) & 1.28\% \\ \midrule

LIME $\cite{Ribeiro2016WhySI}$ & 0.67 (.028) & \multicolumn{1}{l|}{30.6\%} & 0.25 (.005) & \multicolumn{1}{l|}{28.5\%} & 0.86 (.007) & \multicolumn{1}{l|}{17.9\%} & 0.08 (.004) & 
\multicolumn{1}{l|}{32.6\%} & 0.66 (.001) & 22.5\% 
\\ 
TSMULE $\cite{Schlegel2021TSMULELI}$ & -0.42 (.040) & \multicolumn{1}{l|}{1.0\%} & 0.29 (.008) & \multicolumn{1}{l|}{18.9\%} & 0.32 (.029) & \multicolumn{1}{l|}{3.2\%} & 0.08 (.003) & \multicolumn{1}{l|}{4.2\%} & 0.52 (.002) & 33.3\% \\

SHAP $\cite{NIPS2017_7062}$ & -0.32 (.030) & \multicolumn{1}{l|}{2.15\%} & 0.34 (.003) & \multicolumn{1}{l|}{17.7\%} & \textbf{0.92 (.045)} & \multicolumn{1}{l|}{\textbf{15.4\%}} & 0.03 (.013) & \multicolumn{1}{l|}{25.7\%} & 0.55 (.001) & 3.15\% \\ 
TIMESHAP $\cite{bento2020timeshap}$ & -0.34 (.075) & \multicolumn{1}{l|}{34.77\%} & 0.31 (.007) & \multicolumn{1}{l|}{16.8\%} & 0.64 (.037) & \multicolumn{1}{l|}{24.5\%} & 0.02 (.063) & \multicolumn{1}{l|}{5.1\%} & 0.54 (.008) & 3.6\%
\\ 
\midrule
MP  $\cite{Fong2017InterpretableEO}$ & -0.31 (.005) & \multicolumn{1}{l|}{57.9\%} & 0.22 (.001) & \multicolumn{1}{l|}{13.8\%} & 0.14 (.009) & \multicolumn{1}{l|}{63.0\%} & 0.10 (.020) & \multicolumn{1}{l|}{37.9\%} & 0.57 (.021) & 38.9\% \\
DYNAMASK $\cite{Crabbe2021ExplainingTS}$ & -0.35 (.001) & \multicolumn{1}{l|}{1.14\%} & 0.02 (.011) & \multicolumn{1}{l|}{{0.22\%}} & 0.77 (.007) & \multicolumn{1}{l|}{{0.50\%}} & 0.13 (.023) & \multicolumn{1}{l|}{5.43\%} & -0.57 (.031) & {0.30\%} \\
PERT $\cite{Parvatharaju2021LearningSM}$ & 0.73 (.001) & \multicolumn{1}{l|}{25.6\%} & 0.18 (.007) & \multicolumn{1}{l|}{17.7\%} & 0.52 (.070) & \multicolumn{1}{l|}{29.4\%} & -0.02 (.035) & \multicolumn{1}{l|}{38.0\%} & 0.27 (.002) & 26.4\% \\
DEMUX & \textbf{0.81 (.007)} & \multicolumn{1}{l|}{\textbf{0.01\%}} & \textbf{0.42 (.001)} & \multicolumn{1}{l|}{\textbf{9.16\%}} & 0.25 (.004) & \multicolumn{1}{l|}{4.52\%} & \textbf{0.15 (.007)} & \multicolumn{1}{l|}{\textbf{2.43\%}} & \textbf{0.75 (.007)} & \textbf{1.81\%}\\
\bottomrule
\end{tabular}
\caption{Performance of the AUC-Difference $\uparrow$ and IoU $\downarrow$ metrics with the RNN model. Parentheses indicate $\sigma$.}
\label{tab:rnn_mux_auc_table}
\end{table*}

\section{Experimental Study}
\label{sec:exp}
\subsection{Datasets}
\label{sec:Data}
We evaluate our method \textit{DEMUX} on five popular real-world multi-class time series datasets: \textsc{ACSF1 \cite{Gisler2013ACSF1}},
\textsc{Plane \cite{UCRArchive}},
\textsc{Trace \cite{Roverso2000MULTIVARIATETC}}, 
\textsc{Rock \cite{Rockbaldridge}}, 
\textsc{ECG5000 \cite{Goldberger2000PhysioBankPA}}, 
\textsc{Meat \cite{ALJOWDER1997195}}. % DONT BREAK PARAGRAPH HERE -th
Each is a popular  publicly-available dataset \cite{UCRArchive} for multi-class time series classification. The summary statistics are provided in Table \ref{tab:mc_dataset}. We use the default train and test split provided.
For each, we train a three-layered Fully Connected Network (FCN) and a Recurrent Neural Network (RNN) to serve as multi-class deep time series classifiers in need of explanations. Three-layered FCN is considered as a strong baseline in time series classification \cite{Wang2017TimeSC}.
Both models are then trained to achieve state-of-the-art accuracy \cite{UCRArchive} on their respective datasets.
% Our experiments here show the results of both FCN and RNN networks.
% \input{dataset}
% \input{dataset}
% \input{table_dnn_nte_auc}
% \input{table_rnn_nte_auc}

\subsection{Compared Methods}
\label{sec:comp_methods}
We compare our proposed method, DEMUX, to nine state-of-the-art explanation methods. 
% $|P(\hat{y}|X)$

\begin{itemize}

\item \textit{DYNAMASK} \cite{Crabbe2021ExplainingTS}. 
This method is an extension of MP \cite{Fong2017InterpretableEO}
but incorporates the time dependency at each time step for each feature and learns to perturb each time step with static values or adjacent time step values such that $P(\mathbf{\hat{y}}|\mathbf{x})$ decreases. 
Saliency values are then used as the final explanation. 
% DYNAMASK\_DISTINCT applies the DISTINCT \cite{Shimoda2016DistinctCS} method on top of the saliency maps obtained by DYNAMASK.

\item \textit{PERT} \cite{Parvatharaju2021LearningSM}. This method is designed for binary univariate time series and learns to dynamically replace each time step using time step specific interpolation. It learns to perturb each time step such that $P(\mathbf{\hat{y}}|\mathbf{x})$ is preserved to derive instance-specific \textit{for} and \textit{against} evidence. 

\item \textit{TSMULE} \cite{Schlegel2021TSMULELI}. This method is an extension of LIME \cite{Ribeiro2016WhySI} to identify superpixel-like patterns, i.e., semantically related data regions, in time series data. Matrix profile, SAX transformation and uniform segmentation algorithms are used to generate time segments for the perturbation.

\item \textit{TIMESHAP} \cite{bento2020timeshap}. This method builds upon SHAP and extends it to time series to explain important 
%features and 
time steps contributing to the prediction. A temporal coalition pruning method aggregates time events to produce sequences of important data regions.

\item \textit{SHAP} \cite{NIPS2017_7062}. SHAP assigns Shapley values \cite{Charnes1988ExtremalPS} to each time step, thus computing their contributions to $P(\mathbf{\hat{y}}|\mathbf{x})$. 
Each time step in the instance of interest is replaced by a representative value observed at the corresponding time step, sampled from the background dataset.
% \ear{Next sentence makes no sense, as stated.}
% Each time step is replaced by the 
% %every 
% value observed at the corresponding time step across all other instances in the background dataset. 
% In practice, the size of background dataset is reduced due to time constraints.
% SHAP\_DISTINCT applies the DISTINCT \cite{Shimoda2016DistinctCS} method on top of the saliency maps obtained by SHAP.

% \item \textit{TimeSHAP} \cite{bento2020timeshap}. TimeSHAP finds optimal sequence lengths and runs SHAP \cite{NIPS2017_7062} on these sequences to find event-wise and feature-wise saliency maps. 
% TimeSHAP\_DISTINCT applies the DISTINCT \cite{Shimoda2016DistinctCS} method on top of the saliency maps obtained by TimeSHAP.

\item \textit{RISE} \cite{Petsiuk2018RISERI}. The partial derivative of the opaque model's prediction $P(\mathbf{\hat{y}}|\mathbf{x})$ with respect to each time step is estimated empirically by randomly setting time steps to zero and summarizing its impact on the $P(\mathbf{\hat{y}}|\mathbf{x})$. 
% RISE\_DISTINCT applies the DISTINCT \cite{Shimoda2016DistinctCS} method on top of the saliency maps obtained by RISE.

\item \textit{LEFTIST} \cite{Guillem2019AgnosticLE}. The partial derivative of $P(\mathbf{\hat{y}}|\mathbf{x})$ with respect to segments of time steps is estimated empirically by randomly \textit{replacing} the corresponding values from a random instance from the background dataset, or with constants and summarizing its impact on $P(\mathbf{\hat{y}}|\mathbf{x})$. 
% LEFTIST\_DISTINCT applies the DISTINCT \cite{Shimoda2016DistinctCS} method on top of the saliency maps obtained by LEFTIST.

\item \textit{LIME} \cite{Ribeiro2016WhySI}. Saliency values are derived from the coefficients of a linear surrogate model, trained to mimic the opaque model's behaviour in the feature space surrounding $\mathbf{x}$. The approach's success relies on the opaque model behaving linearly locally, which is rarely guaranteed. 
% LIME\_DISTINCT applies the DISTINCT \cite{Shimoda2016DistinctCS} method on top of the saliency maps obtained by LIME.

\item \textit{Meaningful Perturbation (MP)} \cite{Fong2017InterpretableEO}. MP learns to perturb each time step such that $P(\mathbf{\hat{y}}|\mathbf{x})$ decreases. Perturbation is achieved by combining squared exponential smoothing with additive Gaussian noise. Saliency values are then learned iteratively and are used as the final explanation. 
% MP\_DISTINCT applies the DISTINCT \cite{Shimoda2016DistinctCS} method on top of the saliency maps obtained by MP.

\end{itemize}

\subsection{Implementation Details} 
\label{sec:PerformanceEvaluation}
For each dataset, we train a three-layer FCN and a 10-node single-layer RNN with GRU cells to serve as fairly standard multi-class time series classifiers \cite{Wang2017TimeSC} in need of explanations. Each dataset comes with a pre-defined split ratio.
We train each model only on the training data, then explain their predictions for all test instances for each compared explainability method. All reported metrics are the result of the average over five runs to estimate the variance of the explainability methods.
We optimize our proposed method using Adam \cite{kingma2017adam} with a learning rate of $1e^{-3}$ and train for 5000 epochs, which we find empirically achieves convergence. 
We use the Weights \& Biases framework \cite{wandb} for experiment tracking.
Our proposed method is implemented in PyTorch. 
For each compared method, we have used grid search to compute the optimum values of hyperparameters to get the best possible results. Our source code is publicly-available at https://github.com/rameshdoddaiah/DEMUX. This repository includes runtime benchmarks to show the time cost of DEMUX compared to other methods and Hyperparameter tuning table.
% % once
% the paper has been published. % at
% \texttt{hiddenforreview}.

\begin{figure*}[htp!]
    \centering
    % \captionsetup{justification=centering}
    \begin{subfigure}[c]{0.24\textwidth}
    \centering
      \includegraphics[width=1.0\textwidth]{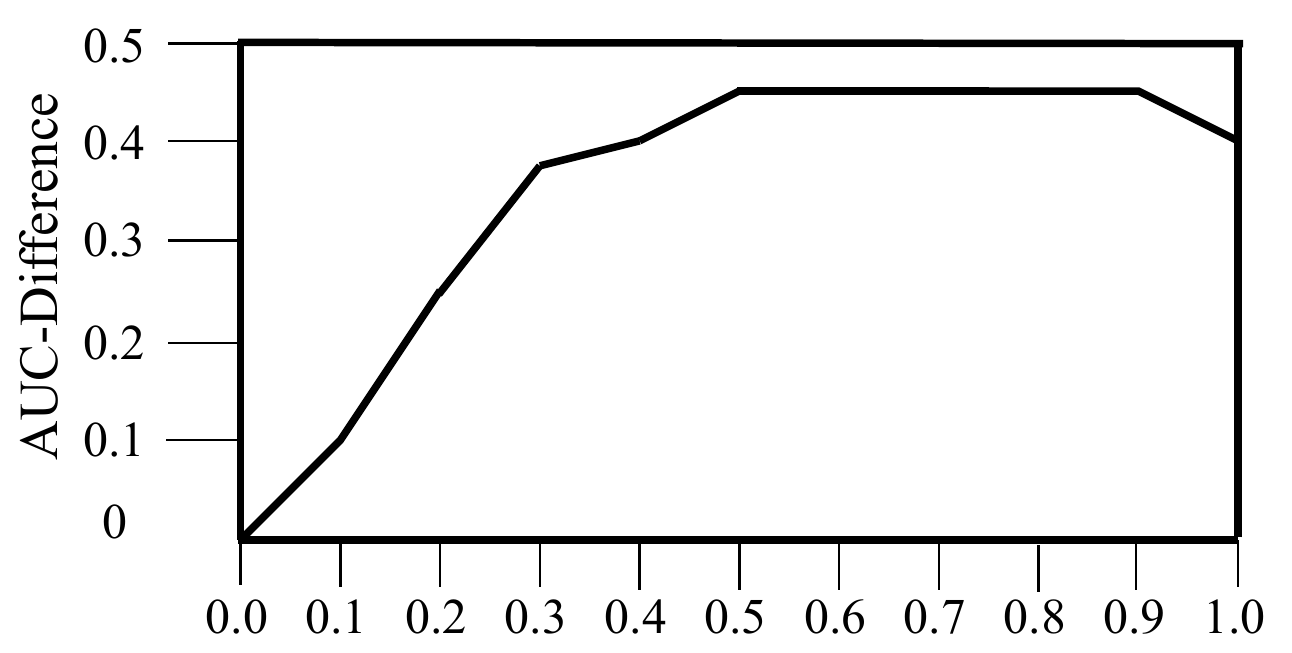}

      \caption{$L_\text{Prev}$ coefficient ($\lambda_1$).}
      \label{fig:LPreservecoeff}
    \end{subfigure}
    \begin{subfigure}[c]{0.24\textwidth}
      \centering
      \includegraphics[width=1.0\textwidth]{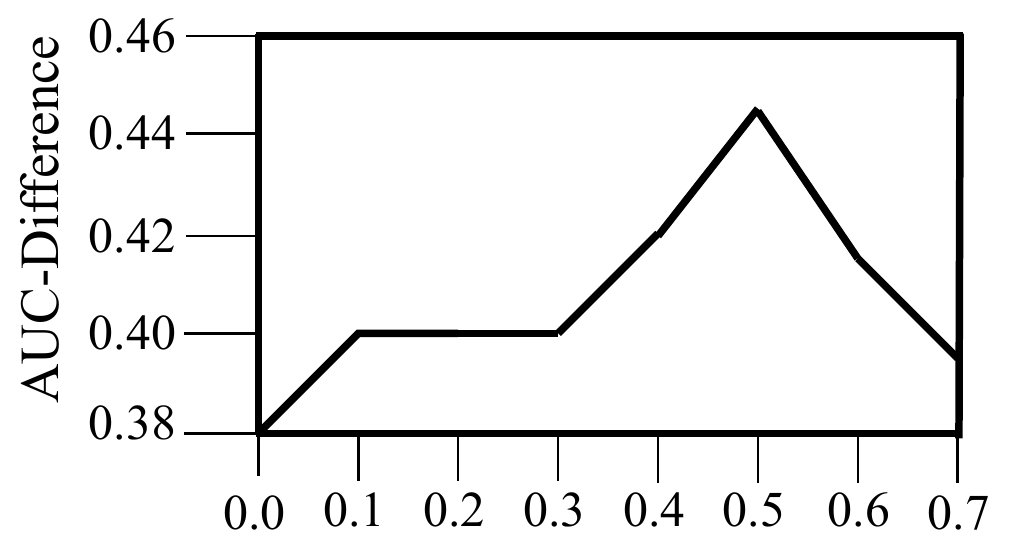}

      \caption{$L_\text{Max}$ coefficient ($\lambda_2$).}
      \label{fig:LMax}
    \end{subfigure}
    \begin{subfigure}[c]{0.24\textwidth}
      \centering
      \includegraphics[width=1.0\textwidth]{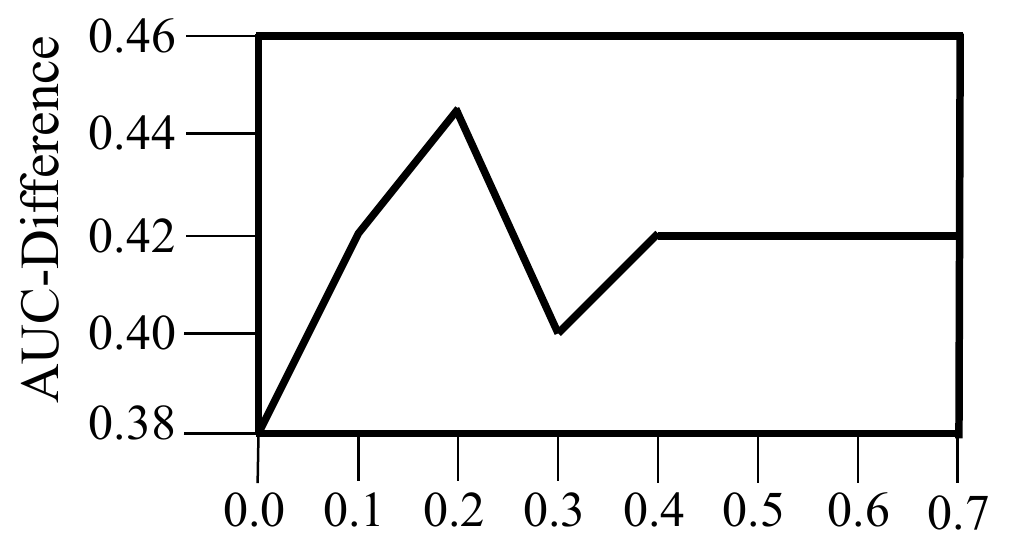}

      \caption{$L_\text{Budget}$ coefficient ($\lambda_3$).}
      \label{fig:LBudget}
    \end{subfigure}
    \begin{subfigure}[c]{0.21\textwidth}
      \centering
      \includegraphics[width=1.0\textwidth]{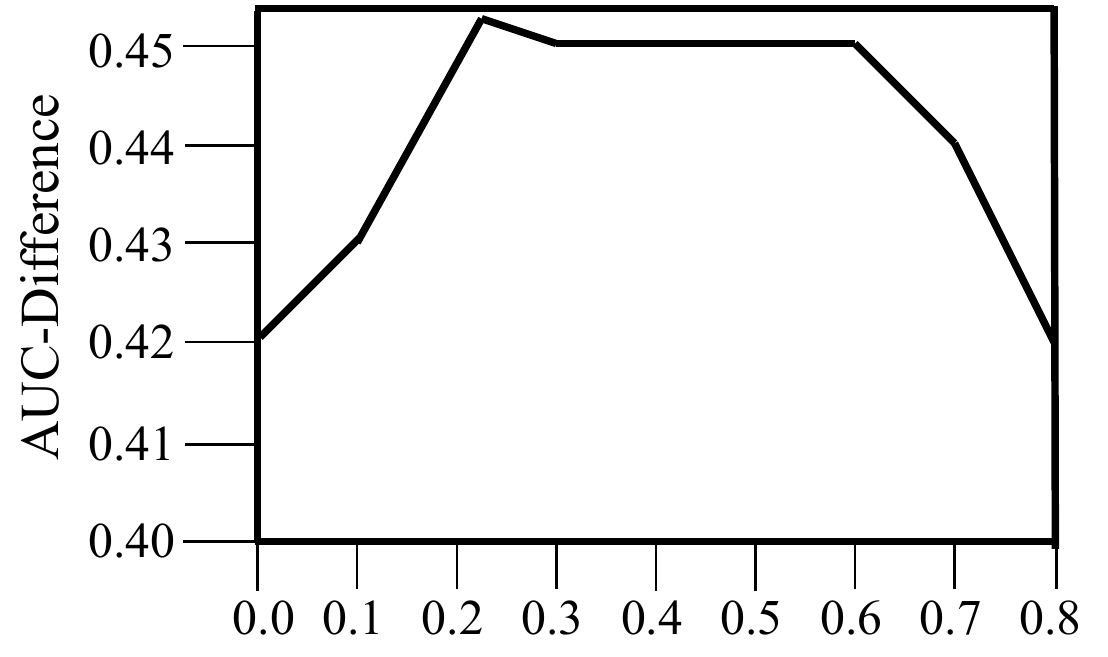}

      \caption{$L_\text{SSD}$ coefficient ($\lambda_5$).}
      \label{fig:LSSDcoeff}
      
    \end{subfigure}

\caption{DEMUX hyperparameter study results on ACSF1 dataset.}

\label{fig:ALLParamsStudy}
% \vspace{-3mm}
\end{figure*}

\begin{figure}[htp!]
 \centering
%   \hspace{-5mm}
    \includegraphics[width=0.46\textwidth]{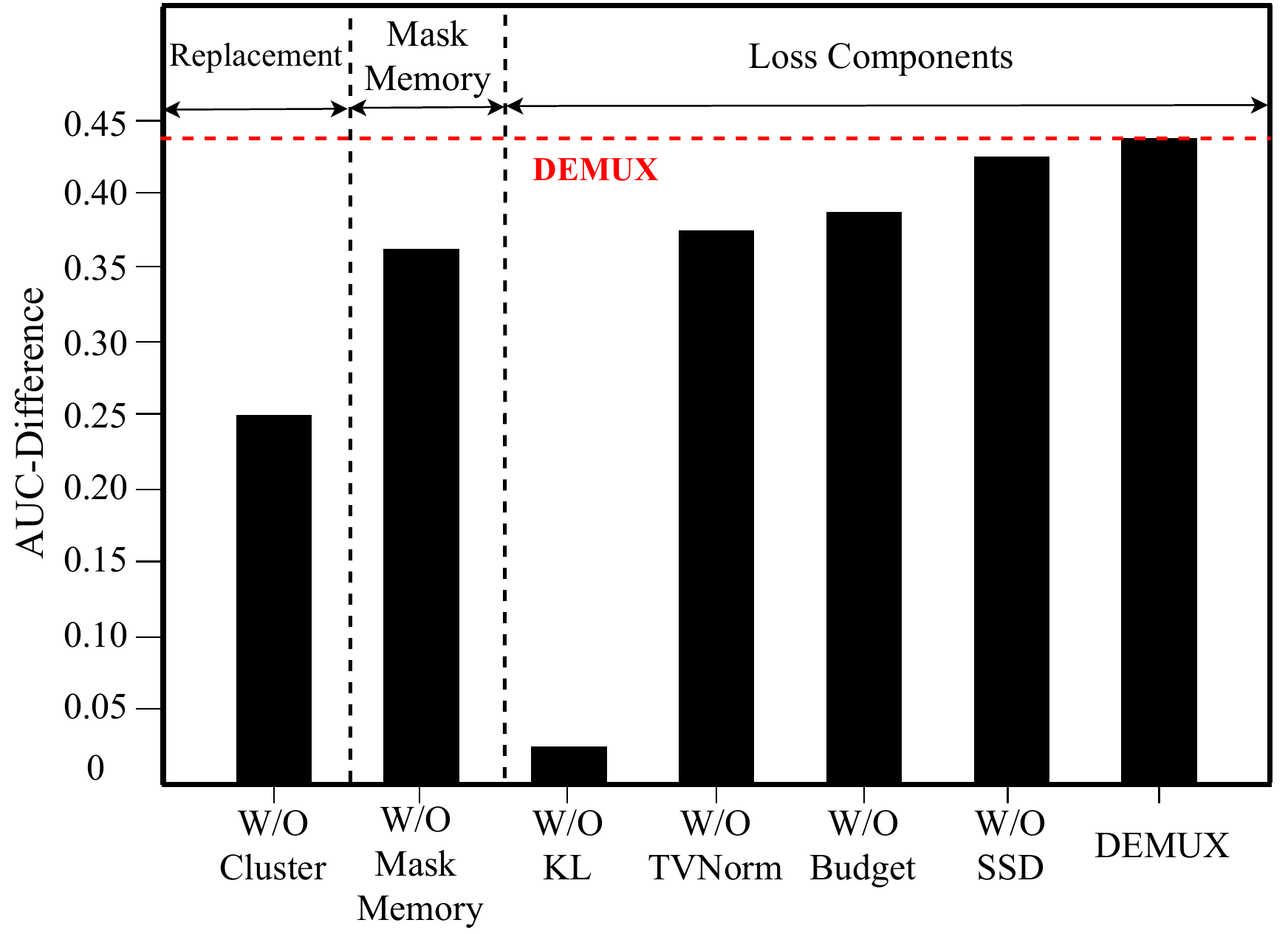}
  \caption{DEMUX ablation study results on ACSF1 dataset.}
  \label{fig:AblationStudy}
\vspace{-4mm}
\end{figure}

\begin{figure*}[t]
 \centering
%   \vspace{-60mm}
    % \includegraphics[width=0.43\textwidth]{fig/LMUX_CaseStudy.pdf}
    % \includegraphics[width=\linewidth]{fig/CIKMCase Study.pdf}
    % \includegraphics[width=\linewidth]{fig/case_study/output2.png}
    
    \includegraphics[width=\linewidth]{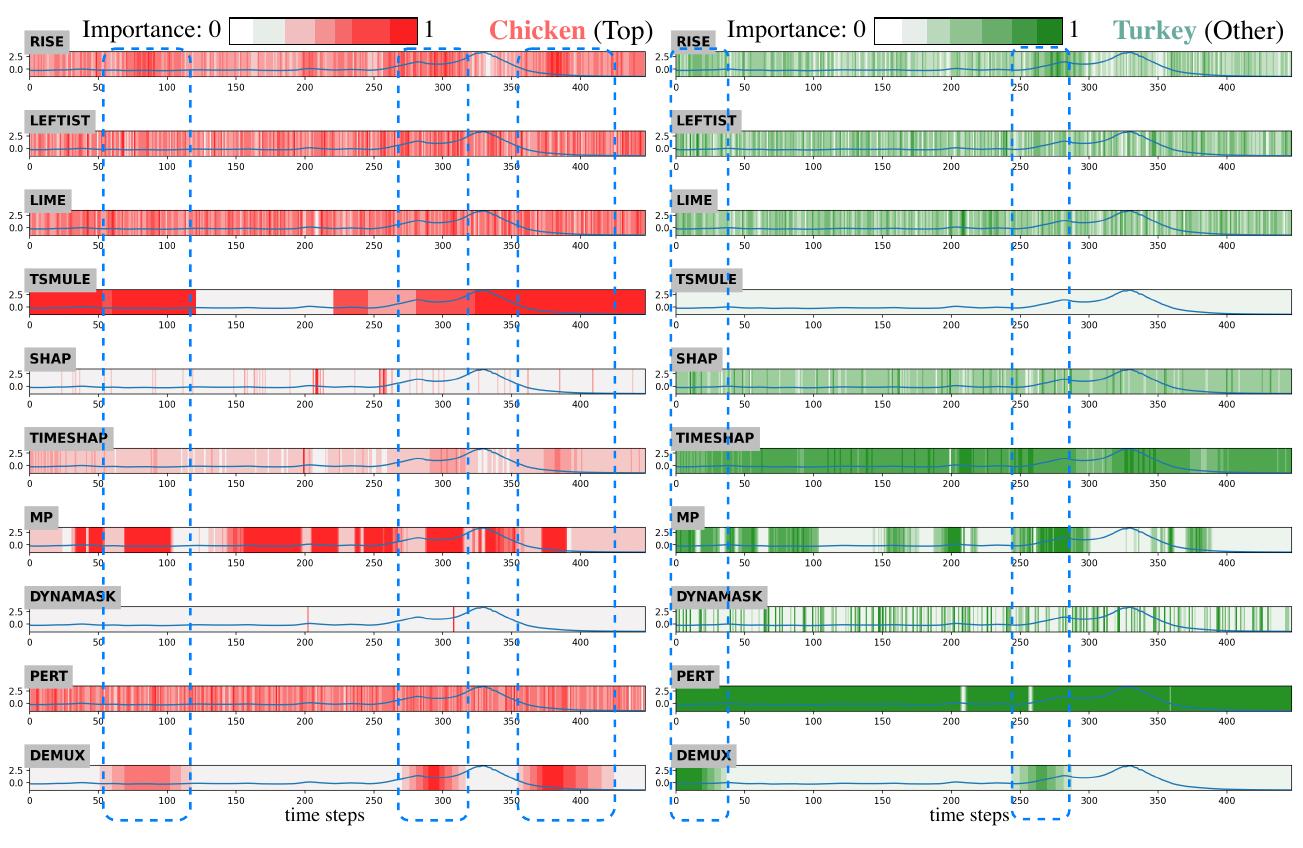}

  \caption{\textsc{Meat \cite{ALJOWDER1997195}} ``Chicken'' Class Case Study. The time series is shown in blue and saliency values are plotted against time steps in {\color{red}red} for the top predicted chicken class and in \textcolor{ForestGreen}{green} for another class, Turkey.
%   XAI methods shown with their relevant saliency maps are RISE, LEFTIST, LIME, TSMULE, SHAP, TIMESHAP, MP, DYNAMASK, PERT and DEMUX.
  DEMUX (the last row) highlights three unique, discriminative subsequences as the class-specific explanation, outlines by blue dashed boxes for the top predicted ``Chicken'' class, unlike the other methods. DEMUX highlights evidence for all classes, but only ``Turkey'' class is shown here due to space constraints.}
  \label{fig:CaseStudy}
% \vspace{-12.5mm}
\end{figure*}

\subsection{Metrics}\label{sec:metrics}
We use two key popular metrics
% \cite{Shimoda2016DistinctCS,Petsiuk2018RISERI,Parvatharaju2021LearningSM}
to evaluate saliency maps for time series under the intuition that a prediction is well-explained if it accurately ranks the class-specific time steps by their importance, as defined by changes in $P(\mathbf{\hat{y}}|\mathbf{x})$, and also returns only the most important time steps and the intersection of 
% \snote{\sout{overlapping} redundant}
overlapping salient regions across the classes are minimal.

% \ear{For AUC - should you also give
% an intuitive name for the metric to reflect
% its purposes in explanation context - maybe something like below:}
% \ear{Measuring Insertion/Deletion Sensitivity Impact:  AuC Metric.} 

\textbf{Measuring Insertion and Deletion Sensitivity Impact: AUC-Difference Metric \cite{hama2022deletion,Petsiuk2018RISERI,Parvatharaju2021LearningSM}.} Saliency maps can be evaluated by ``inserting'' or ``deleting'' time steps from the time series instance based on the derived importance map and observing the changes in the opaque model's predictions \cite{hama2022deletion,Petsiuk2018RISERI,Parvatharaju2021LearningSM}. 
A good saliency map is one that has ranked time steps such that when the most important time steps are inserted or deleted, there is a sharp change in the confidence of the model's prediction. This can be measured by computing the area under the deletion curve (AUDC) as time steps are deleted one by one. A lower value for this area indicates a better explanation. Analogously, insertion of a few important time steps should result in the largest possible increase in the confidence of the model's prediction, thereby creating a large area under the insertion curve (AUIC). To ease comparisons, we merge these two measures into one metric by computing the difference, Equation \ref{auc_diff} represents the difference between AUIC and AUDC where we expect the difference, AUC-Difference should be large (1.0), implying the saliency maps having a high AUIC (1.0) and a low AUDC (0.0) is a good explanation as per PERT \cite{Parvatharaju2021LearningSM,hama2022deletion} and RISE \cite{Petsiuk2018RISERI}. 
% Computing the difference, $\text{AUC-Difference} = \text{AUIC} - \text{AUDC}$ should be large (1.0), implying the saliency maps having a high AUIC (1.0) and a low AUDC (0.0) is a good explanation as per the PERT \cite{Parvatharaju2021LearningSM} and RISE \cite{Petsiuk2018RISERI} methods. 

\begin{equation}
    \label{auc_diff}
    \text{AUC-Difference} = \text{AUIC} - \text{AUDC}
\end{equation}

The metric alleviates the need for human evaluation and
the error-prone collection of human ground-truth.
This   makes it fairer and truer to the classifier’s own view on the problem.
A pivotal choice in insertion and deletion tests is which reference input to use when deleting or inserting values to avoid spurious model classification. 
To achieve deletion during evaluation, we replace a to-be-deleted time step with the farthest cluster's centroid derived from the background data set $D$ to provide in-distribution replacement values. Conversely, for insertion, we start with the the farthest cluster's centroid and iteratively replace the time steps with the values of the instance-of-interest.

% \tnote{do you do this the same way for all compared methods?}
% yes we do the same for all methods for fairer comparison

% \input{table_dnn_mux_auc}
% \input{table_rnn_mux_auc}

% \ear{For IoU - should you also give
% an intuitive name for the metric to reflect
% its purposes - maybe something like below:}
% \ear{Measuring Class-Specificity of an
% Explanation: Intersection over Union (IoU).} 

\textbf{Measuring Class-Specificity of an Explanation: Intersection over Union.} We leverage a standard intersection over union (IoU) metric from the image domain \cite{Shimoda2016DistinctCS}. A simple and a good class-specific saliency map is the one that has no overlapping salient time steps across the classes with respect to the predicted class. This can be measured by intersection over union \cite{Shimoda2016DistinctCS} of salient time steps of the predicted class saliency map with other classes' saliency maps. First, we map each per-class salient time steps to a \textit{boolean value}, either 1 or 0 based on a preset range of thresholds [min=0, max=1, step=0.1]. Second, we find intersection of per-class saliency map with predicted class saliency map and count (IC) number of non-zero salient time steps. Third, we find the union of per-class saliency map with predicted class saliency map and count (UC) number of non-zero salient time steps. 

Equation \ref{iou} represents the ratio of overlapping salient time steps. We repeat step two and three with different thresholds ranging from 0 to 1 at every 0.1 interval to compute area under the curve (IoU curve). Minimal overlapping salient time steps leads to a higher the number of class-specific salient time steps. The area under the curve IoU metric must be minimal to quantitatively say the generated explanation is unique.

\begin{equation}
    \label{iou}
    \text{Intersection over Union ( IoU )} = \frac{IC}{UC}
\end{equation}

\subsection{Experimental Results} 
\subsubsection{DEMUX  finds locally faithful explanations per-class.}\label{local_faithful_experiments}
To address the challenge of generating explanations being locally faithful to the classifier's behavior, we first measure how well all compared explainability methods have \textit{ranked} the time steps by their importance via the AUC-Difference metric \cite{Petsiuk2018RISERI, Parvatharaju2021LearningSM}. % discussed in section \ref{sec:metrics}. 
Our results using Fully-Connected and Recurrent Neural Networks can be found in Tables \ref{tab:fcn_mux_auc_table} and \ref{tab:rnn_mux_auc_table}, respectively, where we evaluate each method on five multi-class datasets.
% DEMUX learns to rank important salient time steps while preserving original prediction probability by the classifier. \tnote{this doesn't say anything about the results?}
In most of the cases, DEMUX significantly outperforms all other methods on AUC-Difference by learning better rankings for time step importance. In general, learning to perturb, as done in DEMUX and PERT, outperforms the random-replacement methods RISE \cite{Petsiuk2018RISERI, Guillem2019AgnosticLE}, LIME \cite{Ribeiro2016WhySI}, and SHAP \cite{bento2020timeshap} for both the long and short time series.
This robust performance demonstrates that learned perturbations effectively produce higher-quality explanations than unlearned perturbations. 

\subsubsection{DEMUX finds class-specific salient time steps.}
\label{cs_measurement}
We next find that most state-of-the-art methods generate per-class saliency maps independently, leading to overlapping salient regions across the classes. As shown in Tables \ref{tab:fcn_mux_auc_table} and \ref{tab:rnn_mux_auc_table}, we also compare all methods according to the IoU metric \cite{Shimoda2016DistinctCS}, which measures how much saliency maps overlap between classes.
As expected, DEMUX largely outperforms existing methods by learning to remove overlaps during training.
Interestingly, DYNAMASK is competitive in three out of the five datasets, though it has much lower AUC-Difference.
Because DYNAMASK seems to sacrifice AUC for IoU, there appears to be a trade-off between these two metrics, as expected.
By and large, DEMUX significantly outperforms the state-of-the-art methods on both metrics, especially when considering them together.

% TODO
% \ear{IS THERE A METRIC that
% combines wins by both metrics; e.g.,
% avg(rank) across both metrics.
% If a method is best (rank1) in one
% metric and 2nd best in 2nd metrics (rank2),
% the its total rank would be  (1+2)/2 =  1.5.
% etc.}

% We next use the IOU metric \cite{Shimoda2016DistinctCS} to measure how well compared methods have addressed this challenge by removing \textit{overlapping} salient regions from the top predicted class to generate unique explanations. Our IOU metric results using Fully Connected Network and Recurrent Neural Network can be found in Tables \ref{tab:fcn_mux_auc_table} and \ref{tab:rnn_mux_auc_table} respectively. In most of the cases, DEMUX significantly outperforms most of the compared methods by successfully learning to remove overlapping salient time steps leading to lower IOU metrics. Unlike compared methods, DEMUX has lower IOU metrics for most of the datasets, it shows DEMUX can produce class-specific explanations with lower overlapping salient regions.

\subsubsection{DEMUX learns consistent explanations.}\label{consistent_experiments}
Some recent works express concern that perturbation methods can lead to different explanations \cite{Kumar2020ProblemsWS} for the same instances when respective saliency masks are initialized randomly.
We share this worry and provide some peace of mind: DEMUX achieves more-consistent explanations across multiple runs than other methods. Each 
saliency mask is re-initialized five times on all time series in each dataset. We report the standard deviation for the AUC-Difference metric as shown in Tables \ref{tab:fcn_mux_auc_table} and \ref{tab:rnn_mux_auc_table}.
DEMUX has a far-lower average standard deviation across all datasets when AUC-Difference and IoU are considered together.

% In general, most of the existing explainability methods produce different explanations for the same time series instance across multiple runs. DEMUX on the other hand consistently produce similar explanations by taking mean of the good saliency maps during learning. The lower standard deviation of AUC-Difference metrics is a measure for this challenge. As seen in tables \ref{tab:fcn_mux_auc_table} and \ref{tab:rnn_mux_auc_table}, we have run each method five times on each dataset to generate the standard deviation. For most of the datasets, our method's AUC-Difference metrics standard deviation is quite low compared to other methods; this shows DEMUX can generate \texit{consistent} explanations across runs.
%We don't have a metrics for this or can we show the standard deviations?
% When time series instances are short, such as for the PLANE and TRACE datasets, random replacement and linear surrogate methods produce comparable results to DEMUX as they rely on a densely-sampled feature spaces surrounding each instance-of-interest. For longer series, covering this space efficiently becomes challenging, motivating \textit{learned class-specific} perturbation.

\label{sec:para}
% Learning an Independent saliency for the top predicted class $L_\text{W/O Distribution Loss}$ without considering its relationship to other classes produces lower AUC-Difference. 

\subsubsection{Ablation study.}\label{sec:ablation}
To demonstrate the need for each component of DEMUX, we perform an ablation study, removing different DEMUX components and reporting the AUC-Difference on the ACSF1 dataset \cite{Gisler2013ACSF1}.
We focus on the replacement strategy, mask memory unit, and each loss component. Our results in Figure \ref{fig:AblationStudy} show that each component is necessary to achieve a good performance.

%RD: Good point professor. We will keep this in mind if needed
% \ear{FIG 4 could be made slightly shorter, if you need to make space.}

\textit{First,} without the per-class time series replacement learning to sample component ($\textit{W/O Cluster}$) AUC-Difference suffers when random time series with irrelevant time steps are used for replacement. This shows the need for selecting class-specific replacement values per time step to create per-class perturbations.
% \tnote{you didn't say what you swapped in place? Do you randomly select? Or do you use PERT's method?}.
\textit{Second,} without the mask memory unit component ($\textit{W/O Mask Memory}$), we pick the final saliency map at $5000^{th}$ epoch as explanation and the AUC-Difference metric decreases slightly. This shows that Perturbation-based method may not consistently pick a good saliency map if stopped after a predefined number of epochs.
% This shows stopping the perturbation-based method after random number of epochs to obtain a saliency map does not consistently provide good explanations.
% \tnote{we don't care HOW it works here... just what the results imply: what did you swap in, what happened (AUC drops), and most importantly, WHAT DOES IT IMPLY?}.  
\textit{Third,} we compare DEMUX's performance while removing different components of the loss function.
As expected, $\text{W/O KL}$ clearly has a substantial impact on the final AUC-Difference, as without this component, there is no relationship between the saliency map and the model's prediction probability distribution. 
\textit{Fourth,} the $\text{W/O TVNorm}$ and $\text{W/O Budget}$ components have less of an impact but still contribute to DEMUX's state-of-the-art performance by learning simple and discriminative subsequences. 
\textit{Fifth,} learning without a class-specific loss component $\text{W/O SSD}$ reduces DEMUX's AUC-Difference further, stressing the need for learning class-specific saliency maps. $\text{DEMUX}$ includes all the above components and indeed achieves the best AUC-Difference.

%weighted KL should go to Hyper parm study
% Second, we experiment with four alternative time series replacement strategies to test the importance of the choice of replacement time series $R$'s impact on opaque box model $f_c$'s prediction $P(C|\hat{X})$. Initially we randomly $L_\text{Random Distance}$ pick an opposite class time series for replacement from the background dataset $\mathcal{D}$. From this experiment, we notice that the AUC-Difference metric is sensitive to the choice of replacement time series $R$. However, by observing the relationship between the predicted confidence of $f_c$ for instance-of-interest $\hat{X}$, $P(C|\hat{X})$ and for the perturbed instance $X^{*}$, $P(C|\hat{X}^{*})$, we propose dual replacement sampling, where we sample a replacement time series from class-of-interest $R_C$ and from opposing class $R_O$, and perform time step specific replacement which results in a substantial improvement in AUC-Difference metric. $L_\text{Random Distance}$ produces a bit lower AUC-Difference as expected, $L_\text{Cluster Distance}$ uses KNN to learn the top N clusters for the given training dataset. It picks the cluster farthest to the given instance to be explained as opposing class and uses prioritization to pick the best replacement time series in this opposing class to outperform all other $L_\text{Mahalanobis}$ and $L_\text{Euclidean}$ replacement strategies.

\subsubsection{Hyperparameter Study.}
Producing class-specific explanations with DEMUX involves balancing the four key hyperparameters: coefficients for $L_\text{SSD}$, $L_\text{Prev}$, $L_\text{Max}$, and $L_\text{Budget}$.
% $L_\text{TVNorm}$,
We investigate the effects of tuning these coefficients in isolation on the ACSF1 dataset and report our results in Figure \ref{fig:ALLParamsStudy}. For each case, we keep all unchanged parameters at their best-found values. 

First, as shown in Figure \ref{fig:LPreservecoeff}, we vary the coefficient $\lambda_{1}$ of $L_\text{Prev}$ 
% \tnote{Doesn't look like Prev in the figure.. it says SSD... oh you means preservation? talk about them in the same order as the figure always!!!!} 
between 0 to 1. This  maintains the prediction probability distribution across all the classes and we find a value of 0.7 suffices. A low value for $\lambda_1$ fails to preserve the model's probability distribution, so the loss function may not encourage instance-specific explanations. There, we notice a gradual increase in AUC-Difference until it stabilizes before dropping for higher values of $\lambda_{1}$.
% \tnote{we don't need to say what we see precisely, just that it goes up, then down, which indicates the best value is somewhere in the middle}.
% Too high, and this component could overpower other loss components. 
Second, a low value for the coefficient of $L_\text{Max}$ $\lambda_2$  does not generate good explanations for low probability predictions. Here a high value discards salient time steps from the explanations as shown in Figure \ref{fig:LMax}. AUC-Difference increases until $\lambda_2$ reaches 0.5 and decreases for values $\ge 0.5$.
Third, too-low $L_\text{Budget}$ coefficient produces redundant salient time steps, while too high removes important salient time step. AUC-Difference gradually climbs up until this coefficient reaches a value of 0.2, and then it starts decreasing and later stabilizes. The optimal value of 0.2 provides a simple and meaningful explanation as shown in Figure \ref{fig:LBudget}. 
Fourth, we vary the coefficients tween 0 to 0.8 and find the optimal value is 0.2 as shown in Figure \ref{fig:LSSDcoeff}. Too low means not enough focus on class-specificity, while a high value could remove important regions in favor of class-specificity. 
% \tnote{these discuss what could happen when you vary them, not what actually happens.. you should try and comment on what you actually see, too. Like $L_max$ goes up then back down...}

\subsubsection{Case Study on the \textsc{Meat \cite{ALJOWDER1997195}} dataset.}
\label{sec:case_study_mc}
Finally, we describe a case study conducted using DEMUX to explain an RNN's predictions for an instance from the \textsc{Meat} dataset \cite{ALJOWDER1997195}. Each instance of \textsc{Meat} dataset has 500 time steps. This is a multi-class classification task: \textsc{Meat} is a food spectrograph dataset describing food types as class---a crucial task in food safety and quality assurance. There are three
classes: \textit{Chicken}, \textit{Pork}, and \textit{Turkey}, which are predicted with probabilities 88\%, 11\%, and 1\%, respectively for the  instance we consider.

Our case study results are shown in Figure \ref{fig:CaseStudy}, where each method's time step importance is \textbf{{\color{red}red}} for class \textit{Chicken} and \textbf{\textcolor{ForestGreen}{green}} for \textit{Turkey}, with darker hues indicating higher importance.
By visually investigating discriminative subsequences, we  see that the compared methods highlight several regions of input time series as important for the top predicted class \textit{Chicken}.
On the contrary, DEMUX (the last row) succeeds to highlight the discriminative subsequences as class-specific evidence for the top predicted class \textit{Chicken}. Unlike the compared methods, DEMUX has reduced overlapping salient regions between the top predicted class \textit{Chicken} and class \textit{Turkey} as shown in the blue dashed vertical boxes (\textit{Pork} class is not shown due to space constraints). These results suggest that DEMUX's superiority over state-of-the-art methods is due to the ability to learn class-specific evidence with respect to other classes. 
\section{Conclusion}
\label{sec:conclus}
With this work, we identify class-specificity as a 
critical criteria for meaningful  explanations of   multi-class deep time series classifiers.
Solving this open problem is essential for users who seek explanations for why a model predicted one class \textit{in particular} \cite{Shimoda2016DistinctCS}.
We introduce DEMUX, the first solution to this open problem, which extends beyond recent methods for perturbation-based explainability for time series models by learning to encourage class-specificity during training via a novel loss function.
With three interdependent modules, all of which are learned, DEMUX successfully produces class-specific explanations.
In our experiments, we compare DEMUX to nine state-of-the-art explainability methods for two 
%opaque
deep models on five multi-class datasets.
Our results demonstrate that DEMUX's explanations are (1) more class-specific than the alternatives, which leads to (2) higher quality  according to popular explainability metrics, and (3) consistent explanations across multiple initializations.

% Using a case study, we also demonstrate that DEMUX's explanations are easy to understand and useful to the end user.

% In this work, we identify the need for attribution-based class-specific explanations for deep multiclass time series classifiers. We then design DEMUX, a method that learns to highlight the class-specific salient time steps as evidence and the degree to which they are important for the classifier's top prediction.
% By adaptively \textit{learning} a perturbation and shared saliency removal function, DEMUX creates \textit{unique} and \textit{consistent} explanations for the top predicted class. This evidence, presented as one unique importance value per time step, thus allows an end-user to understand what information their classifier was using during multiclass prediction, bringing much needed clarity and transparency in deep learning time series models internals.
% In our experiments and case study, we conclusively demonstrate that DEMUX accurately discovers the class-specific salient time steps as it outperforms seven state-of-the-art alternatives on two key metrics on five datasets.

% balance the two columns of the last page
\balance

% set the style of references
\bibliographystyle{ieeetr}
\bibliography{main.bib}

\begin{thebibliography}{10}

\bibitem{Dixon2020FinancialFW}
M.~F.~D. et~al., ``Financial forecasting with $\alpha$-rnns: A time series
  modeling approach,'' in {\em Frontiers in Applied Mathematics and
  Statistics}, 2020.

\bibitem{Fulcher2017hctsaAC}
B.~D.~F. et~al., ``hctsa: A computational framework for automated time-series
  phenotyping using massive feature extraction.,'' {\em Cell systems}, 2017.

\bibitem{Che2018RecurrentNN}
Z.~Che and et~al., ``Recurrent neural networks for multivariate time series
  with missing values,'' {\em Scientific Reports}, 2018.

\bibitem{Li2021ShapeNetAS}
G.~L. et~al., ``Shapenet: A shapelet-neural network approach for multivariate
  time series classification,'' in {\em AAAI}, 2021.

\bibitem{tonekaboni2020went}
Tonekaboni and et~al., ``What went wrong and when? instance-wise feature
  importance for time-series models,'' {\em NeurIPS}, 2020.

\bibitem{Schlegel2021TSMULELI}
U.~S. et~al., ``Ts-mule: Local interpretable model-agnostic explanations for
  time series forecast models,'' in {\em PKDD/ECML Workshops}, 2021.

\bibitem{Parvatharaju2021LearningSM}
P.~S. Parvatharaju, R.~Doddaiah, T.~Hartvigsen, and E.~A. Rundensteiner,
  ``Learning saliency maps to explain deep time series classifiers,'' {\em
  Proceedings of the 30th ACM International Conference on Information \&
  Knowledge Management}, 2021.

\bibitem{Crabbe2021ExplainingTS}
J.~Crabbe and M.~van~der Schaar, ``Explaining time series predictions with
  dynamic masks,'' in {\em ICML}, 2021.

\bibitem{bento2020timeshap}
J.~e.~a. Bento, ``Timeshap: Explaining recurrent models through sequence
  perturbations,'' {\em ArXiv}, 2020.

\bibitem{Guidotti2020ExplainingAT}
R.~Guidotti and et~al., ``Explaining any time series classifier,'' {\em
  International Conference on Cognitive Machine Intelligence}, 2020.

\bibitem{Guillem2019AgnosticLE}
M.~Guillem{\'e} and et~al., ``Agnostic local explanation for time series
  classification,'' {\em ICTAI}, 2019.

\bibitem{Ribeiro2016WhySI}
M.~T. Ribeiro, S.~Singh, and C.~Guestrin, ``"why should i trust you?":
  Explaining the predictions of any classifier,'' {\em CoRR}, 2016.

\bibitem{NIPS2017_7062}
L.~et.al, ``A unified approach to interpreting model predictions,'' {\em
  Advances in Neural Information Processing Systems 30}, 2017.

\bibitem{ismail2020benchmarking}
A.~A. Ismail, M.~Gunady, H.~Bravo, and S.~Feizi, ``Benchmarking deep learning
  interpretability in time series predictions,'' 2020.

\bibitem{Shimoda2016DistinctCS}
W.~Shimoda and K.~Yanai, ``Distinct class-specific saliency maps for weakly
  supervised semantic segmentation,'' in {\em ECCV}, 2016.

\bibitem{Kaur2020InterpretingIU}
H.~Kaur and et. al, ``Interpreting interpretability: Understanding data
  scientists' use of interpretability tools for machine learning,'' {\em CHI
  Conference}, 2020.

\bibitem{Ribeiro_2020}
Ribeiro and et~al., ``Automatic diagnosis of the 12-lead ecg using a deep
  neural network,'' {\em Nature Communications}, 2020.

\bibitem{mujkanovic2020timexplain}
F.~e.~a. Mujkanovic, ``timexplain--a framework for explaining the predictions
  of time series classifiers,'' {\em ArXiv}, 2020.

\bibitem{Fong2017InterpretableEO}
R.~Fong and A.~Vedaldi, ``Interpretable explanations of black boxes by
  meaningful perturbation,'' {\em ICCV}, 2017.

\bibitem{Kumar2020ProblemsWS}
I.~Elizabeth and K.~et~al., ``Problems with shapley-value-based explanations as
  feature importance measures,'' in {\em ICML}, 2020.

\bibitem{fong2019understanding}
R.~Fong, M.~Patrick, and A.~Vedaldi, ``Understanding deep networks via extremal
  perturbations and smooth masks,'' in {\em ICCV}, 2019.

\bibitem{Kelany2020DeepLM}
Omnia and et~al., ``Deep learning model for financial time series prediction,''
  {\em International Conference on Innovation in Information Technology}, 2020.

\bibitem{Mehdiyev2017TimeSC}
N.~Mehdiyev and et~al., ``Time series classification using deep learning for
  process planning: A case from the process industry,'' {\em Procedia Computer
  Science}, 2017.

\bibitem{Zhang2018HumanAR}
Y.~Zhang and et~al., ``Human activity recognition based on time series analysis
  using u-net,'' {\em ArXiv}, 2018.

\bibitem{schlegel2019rigorous}
U.~S and et~al., ``Towards a rigorous evaluation of xai methods on time
  series,'' {\em CoRR}, 2019.

\bibitem{Satopaa2011FindingA}
V.~Satopaa and et~al., ``Finding a "kneedle" in a haystack: Detecting knee
  points in system behavior,'' {\em ICDCS}, 2011.

\bibitem{Schaul2016PrioritizedER}
T.~Schaul, J.~Quan, I.~Antonoglou, and D.~Silver, ``Prioritized experience
  replay,'' {\em CoRR}, vol.~abs/1511.05952, 2016.

\bibitem{Petsiuk2018RISERI}
V.~Petsiuk, A.~Das, and K.~Saenko, ``Rise: Randomized input sampling for
  explanation of black-box models,'' {\em CoRR}, vol.~abs/1806.07421, 2018.

\bibitem{Gisler2013ACSF1}
Gisler and et~al., ``Appliance consumption signature database and recognition
  test protocols,'' in {\em WoSSPA}, 2013.

\bibitem{UCRArchive}
Y.~Chen, E.~Keogh, B.~Hu, N.~Begum, A.~Bagnall, A.~Mueen, and G.~Batista, ``The
  ucr time series classification archive,'' 2015.
\newblock \url{www.cs.ucr.edu/~eamonn/time_series_data/}.

\bibitem{Roverso2000MULTIVARIATETC}
D.~Roverso, ``Multivariate temporal classification by windowed wavelet
  decomposition and recurrent neural networks,'' 2000.

\bibitem{Rockbaldridge}
A.~Baldridge, S.~Hook, C.~Grove, and G.~Rivera, ``The aster spectral library
  version 2.0,'' {\em Remote Sensing of Environment}, 2009.

\bibitem{Goldberger2000PhysioBankPA}
A.~L. Goldberger and et~al., ``Physiobank, physiotoolkit, and physionet:
  components of a new research resource for complex physiologic signals.,''
  {\em Circulation}, 2000.

\bibitem{ALJOWDER1997195}
O.~Al-Jowder, E.~Kemsley, and R.~Wilson, ``Mid-infrared spectroscopy and
  authenticity problems in selected meats,'' {\em Food Chemistry}, 1997.

\bibitem{Wang2017TimeSC}
Z.~Wang and et~al., ``Time series classification from scratch with deep neural
  networks: A strong baseline,'' {\em IJCNN}, 2017.

\bibitem{Charnes1988ExtremalPS}
A.~Charnes and et~al., ``Extremal principle solutions of games in
  characteristic function form: Core, chebychev and shapley value
  generalizations,'' Springer Netherlands, 1988.

\bibitem{kingma2017adam}
D.~P. Kingma and J.~Ba, ``Adam: A method for stochastic optimization,'' {\em
  CoRR}, vol.~abs/1412.6980, 2015.

\bibitem{wandb}
L.~Biewald, ``Experiment tracking with weights and biases,'' 2020.
\newblock Software available from wandb.com.

\bibitem{hama2022deletion}
N.~Hama, M.~Mase, and A.~B. Owen, ``Deletion and insertion tests in regression
  models,'' 2022.

\end{thebibliography}

\end{document}